\documentclass{article}

\usepackage[final]{neurips_2024}

\usepackage[utf8]{inputenc} 
\usepackage[T1]{fontenc}    
\usepackage{hyperref}       
\usepackage{url}            
\usepackage{booktabs}       
\usepackage{amsfonts}       
\usepackage{nicefrac}       
\usepackage{microtype}      
\usepackage{xcolor}         
\usepackage{graphicx}
\usepackage{amsmath}
\usepackage{color}
\usepackage{lipsum}
\usepackage{wrapfig}

\title{How Diffusion Models Learn to Factorize and Compose}

\author{%
Qiyao Liang \quad Ziming Liu \quad Mitchell Ostrow \quad Ila Fiete \\
Massachusetts Institute of Technology\\
\texttt{\{qiyao,zmliu,ostrow,fiete\}@mit.edu}\\
}

\begin{document}

\maketitle

\begin{abstract}
Diffusion models are capable of generating photo-realistic images that combine elements which likely do not appear together in the training set, demonstrating the ability to \textit{compositionally generalize}. Nonetheless, the precise mechanism of compositionality and how it is acquired through training remains elusive. Inspired by cognitive neuroscientific approaches, we consider a highly reduced setting to examine whether and when diffusion models learn semantically meaningful and factorized representations of composable features. We performed extensive controlled experiments on conditional Denoising Diffusion Probabilistic Models (DDPMs) trained to generate various forms of 2D Gaussian bump images. We found that the models learn factorized but not fully continuous manifold representations for encoding continuous features of variation underlying the data. With such representations, models demonstrate superior feature compositionality but limited ability to interpolate over unseen values of a given feature. Our experimental results further demonstrate that diffusion models can attain compositionality with few compositional examples, suggesting a more efficient way to train DDPMs. Finally, we connect manifold formation in diffusion models to percolation theory in physics, offering insight into the sudden onset of factorized representation learning. Our thorough toy experiments thus contribute a deeper understanding of how diffusion models capture compositional structure in data.    
\end{abstract}


\section{Introduction}
\label{sec:intro}
Large-scale text-to-image generative models can produce photo-realistic synthetic images that combine elements in a novel fashion (compositional generalization). Nonetheless, the ability of models to do so, as well as their failure modes, are not well-studied systematically in large diffusion models due to the size of the model and the complex and high-dimensional nature of their training datasets. Factorization and compositional generalization have been theoretically and empirically investigated in many deep generative models before~\cite{zhao2018, beta_vae, bvae_disentanglement, role_of_disentanglement, xu2022, okawa2023, first_principle, factorization_ANN, emergent_language}. However, these studies have not reached a unanimous conclusion on whether factorized representations learned in the intermediate layers of the model promote compositional generalization in the model performance. Specifically, several studies~\cite{xu2022, role_of_disentanglement, emergent_language} have found little correlation between factorization and compositionality, contrary to others that suggest that factorization promotes compositionality~\cite{bengio_factorization, beta_vae, bvae_disentanglement, duan, factorization_ANN}. As a result, there is no consensus on the exact mechanism of compositionality and how models gain the ability to compositionally generalize. However, these previous studies involve data with a complex mix of discrete and continuously varying features, making it difficult to explicitly analyze the model's learned representations beyond the disentanglement score, and thereby hindering a deeper understanding of factorization and compositionality.

Despite the remarkable compositional abilities demonstrated by diffusion-based text-to-image generative models, explicit exploration of factorization and compositional generalization of diffusion models has been relatively limited. In a recent empirical study of toy diffusion models, Okawa et. al.~\cite{okawa2023} shows that diffusion models learn to compositionally generalize multiplicatively. They too, however, did not focus on the mechanistic aspect of compositional generalization within the model or analyze the representations learned by the models. To investigate factorization and compositionality in diffusion models, we conduct extensive experiments on conditional denoising diffusion probablistic models (DDPMs) \cite{ho2020denoising} trained with various 2D Gaussian datasets to characterize their learned representations and generalization performance.\footnote{A preliminary version of this work was presented at the ME-FoMo and BGPT workshops at ICLR 2024 \cite{diff_liang}. } Drawing inspiration from traditional cognitive psychology and neuroscience studies on animals and humans, we design straightforward cognitive tasks for models to perform while analyzing both their generation output (``behavior'') and their intermediate layer representations (``neural activities''). Specifically, we evaluate the diffusion model's ability to generalize by generating 2D Gaussian bumps centered at specified $x,y$ coordinates on a finite canvas. We deliberately chose a low-dimensional dataset and a simplified experimental setup to enable a clear analysis of how the model's representation factorization relates to its generalization capabilities. 

Conceptually, the models can adopt different parameterizations of the $x,y$ coordinates for the Gaussian bumps, each with varying implications for data efficiency. Consider a $K$-dimensional dataset created from the composition of multiple independently varying 1-dimensional (1D) latent features (e.g. a factorizable data distribution $P(x_1, x_2, \ldots, x_K) = P(x_1)P(x_2)\ldots P(x_K)$). Hypothetically, a model may learn either a joint representation of all $K$ dimensions $P(x_1, x_2, \ldots, x_K) \neq P(x_1)P(x_2)\ldots P(x_K)$ or learn independent 1D representations $P(x_1)P(x_2)\ldots P(x_K)$. In the former case, if there were $N$ states per dimension, the model would need to see approximately $N^K$ samples during training. If the model could somehow recognize the independence of the different factors, as few as $KN$ training samples might suffice. We term the former a \textit{coupled} representation, while the latter is a \textit{factorized} one. Factorized representations could be far more sample efficient to learn, potentially permitting better generalization, and are more robust and adaptable for downstream applications. In biological neural networks, features of independent variation are typically represented in a factorized manner. For instance, primates encode allocentric pose information in a factorized fashion, leveraging grid cells for two-dimensional location and head direction cells for orientation \cite{factorization_biology, factorization_biology2, grid, Chaudhuri2019-yh, Taube1990-vf, Taube2007-if, Hafting2005-nk, Stensola2012-nz, Burak2009-rk}. Further, spatial information about objects in the environment is factorized into place cells and other spatial coding cells~\cite{Moser_Rowland_Moser_2015, factorization_biology2}. 

Across three related studies, we investigate how diffusion models learn to encode independent features of the dataset, how that informs the model's ability to generalize out-of-training-distribution in terms of composition and interpolation, and how variations in the training data could impact such generalization. To be precise, here we define \textit{compositionality} as the ability to generate novel combinations of feature values that are not observed during training (e.g., given $\{(a, b), (c, d)\}$, output $\{(a, d), (c, b)\}$). We define \textit{interpolation} as the ability to linearly combine different feature values (e.g., output $x=16$ given $x=15$ and $x=17$). The failure of the model to perform the designed tasks could be due to its inability to handle composition, interpolation, or a combination of both. We train the model on 2D Gaussian data and, in some cases, explicitly include 1D Gaussian stripes with specified $x,y$ positions that can be combined to form the desired 2D Gaussian. We then pose the following questions:
\begin{enumerate}
    \item Do diffusion models learn factorized representations? If so, when do these emerge over training? 
    \item Do the trained models learn to generalize beyond the training distribution, and what kind of training data is sufficient for this generalization?
    \item Does the inclusion of a few explicitly factorized examples in training (e.g. the 1D Gaussian data) improve sample efficiency and generalization? 
\end{enumerate}

\textbf{Contributions.} In three experiments, we find that:
\begin{enumerate}
    \item Given factorized, continuous conditional inputs, diffusion models learn ``hyper-factorized'' representations that are orthogonal not only for independent features but also different values of the same feature (Section~\ref{sec:res-factorization}). 
    \item Models compose well but interpolate poorly. Models can compositionally generalize when they observe the full extent of each independent latent feature along with a few compositional examples (Section~\ref{sec:res-composition}). 
    \item Models trained on datasets containing isolated factors of variation require an exceptionally small number of compositional examples to compositionally generalize, showing remarkable data efficiency (Section~\ref{sec:res-composition}).
    \item Formation of a factorized representation for each continuous dimension of variation requires a threshold amount of correlated data, related to percolation theory in physics (Section~\ref{sec:res-perc}).
\end{enumerate}
Our results suggest that diffusion models can learn factorized, though not fully continuous, representations of continuous features with independent variations. These models demonstrate exceptional compositionality but have limited interpolation ability, even with sufficient training and data. Our analysis offers deeper diagnostic insights into the mechanism of factorization and compositionality of diffusion models from a microscopic representation perspective. Moreover, our insights on the data efficiency of training with isolated factors of variation with few compositional examples suggest the possibility of far more data efficient methods of training diffusion models.

\section{Methods}
\label{sec:setup}
\begin{wrapfigure}{r}{0.35\textwidth}
\begin{center}
\includegraphics[width=0.35\textwidth]{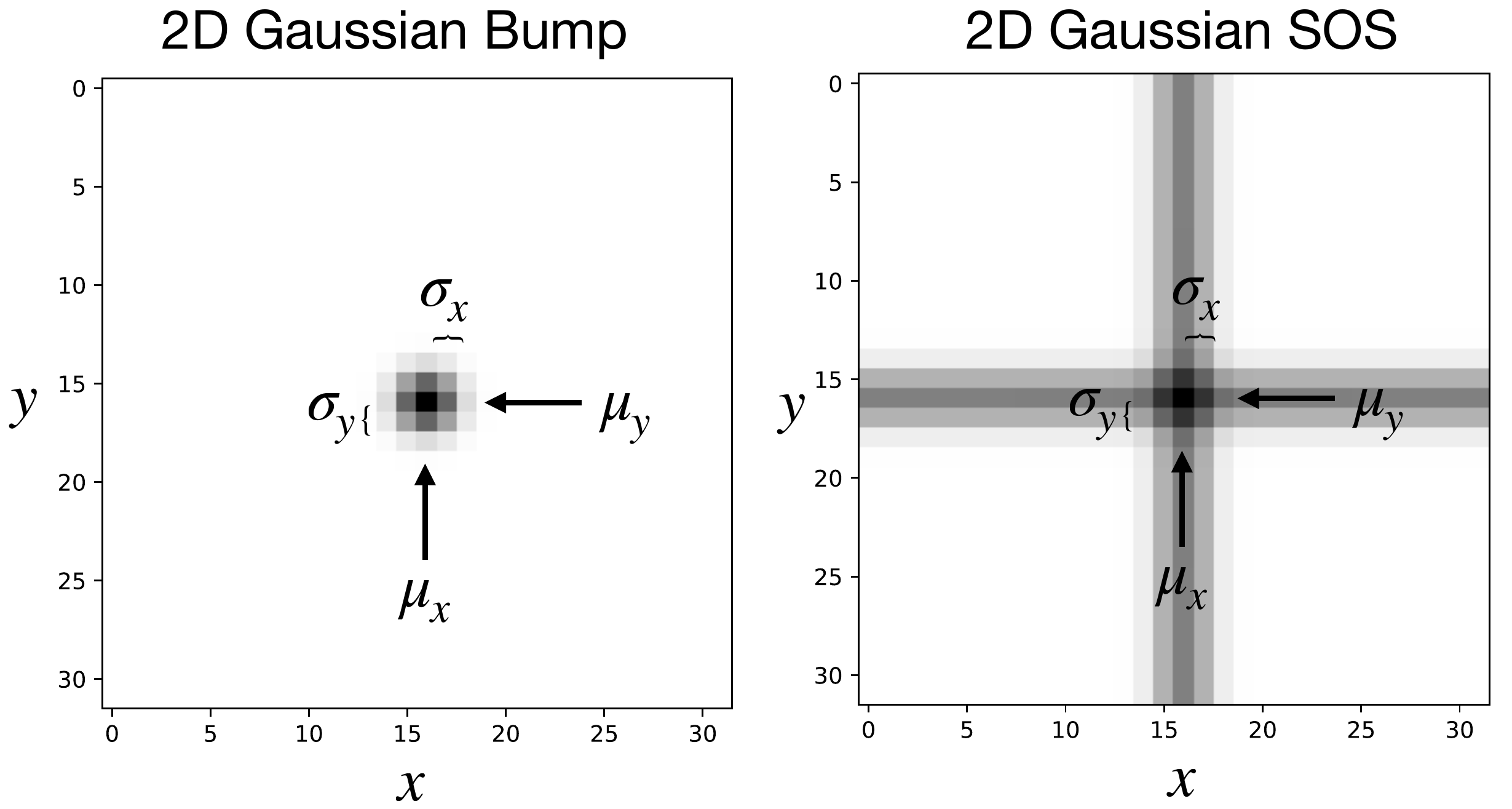}
\end{center}
\caption{Example $32\times 32$ image data of a 2D Gaussian bump (left) and a 2D Gaussian SOS (right).}
\label{fig:sample_image}
\end{wrapfigure}
\textbf{Dataset. }
We generate $N \times N$ pixel grayscale images. By default, we set $N=32$ unless otherwise specified. Each image contains one 2D Gaussian bump (``blob''), the multiplication of a vertical and a horizontal 1D Gaussian stripe, or one 2D Gaussian addition, the sum of a vertical and a horizontal 1D Gaussian stripe (``sum of stripes (SOS's)''), at various $x,y$ locations.
The brightness $v_{(x,y)}$ of the pixel at position $(x,y)$ for a 2D Gaussian blob centered at $(\mu_x, \mu_y)$ with standard deviation $(\sigma_x, \sigma_y)$ is given as  $v^{\text{bump}}_{(x,y)} = 255 \times (1-e^{-(x-\mu_x)^2/4\sigma_x^2-(y-\mu_y)^2/4\sigma_y^2})$  and as $v^{\text{SOS}}_{(x,y)} = 255 \times [1-\frac{1}{2}(e^{-(x-\mu_x)^2/4\sigma_x^2}+e^{-(y-\mu_y)^2/4\sigma_y^2})]$ for a 2D Gaussian SOS with the normalized range of $v_{(x,y)}$ to be $[0,255]$. Each image is generated with a ground truth label of $(\mu_x, \mu_y)$, which continuously vary within $[0,N]^2$ unless otherwise specified. In our convention of notation, we label the top left corner of the image as $(1,1)$ while the bottom right corner of the image as $(N,N)$. Sample $32\times 32$ images of 2D Gaussian bump and SOS centered at $\mu_x=\mu_y=16$ with $\sigma_x=\sigma_y=1$ are shown in Fig.~\ref{fig:sample_image}.

A single dataset of these images consist of the enumeration of all possible Gaussians tiling the whole $N \times N$ canvas at increment $d_x$ in the $x$-direction and $d_y$ in the $y$-direction. A larger $d_x$ or $d_y$ means a sparser tiling of the image space and fewer data while a smaller $d_x$ or $d_y$ result in more data with denser tiling of the total image space. Moreover, each 2D Gaussian have independent spread in the $x$ and $y$ direction given by $\sigma_x$ and $\sigma_y$, with a larger spread leading to more spatial overlap of neighboring Gaussians and a smaller spread less overlap. By parameterically tuning the increments $d_x$ and $d_y$ and the spread $\sigma_x$ and $\sigma_y$, we can generate datasets of various sparsities and overlaps. We provide a more detailed analysis of the various attributes of the data based on these parameters in Appendix~\ref{sec:app_dataset}. For the majority of the experimental results we show in Sec.~\ref{sec:results}, we have chosen to fix $\sigma:=\sigma_x=\sigma_y=1.0$ and $d:=d_x=d_y=0.1$ unless otherwise specified. A default $32\times 32$ dataset of $d=0.1$ contains 102400 images.

\textbf{Models \& Evaluations. } 
We train a conditional DDPM~\citep{ho2020denoising,cond1,cond2} with a standard UNet architecture as shown in Appendix Fig.~\ref{fig:architecture}. For each image in the training dataset, we provide an explicit ground truth label $(\mu_x,\mu_y)$ as an input to the network. For reference, we investigate the internal representation learned by the model using the output of layer 4 as labeled in Fig.~\ref{fig:architecture}. Since each dataset has inherently two latent dimensions, $x$ and $y$, we use dimension reduction tools such as PCA or UMAP~\cite{mcinnes2020umap} to reduce the internal representation of the model to a 2D/3D embedding for the ease of visualization and analysis.
We defer the details of model architecture, dimension reduction, and experimentation to Appendix~\ref{sec:app_architecture} and ~\ref{sec:app_umap}. Briefly, we assess the performance of the model based on the accuracy of its generated images at correctly displaying the center location of the 2D Gaussian. Further details on the precise definition of the accuracy metric can be found in Appendix~\ref{sec:app_metrics}.

\section{Results}
\label{sec:results}
\subsection{Models learn factorized but not fully continuous representations}
\label{sec:res-factorization}

\begin{figure}[!htb]
\begin{center}
\includegraphics[width=5in]{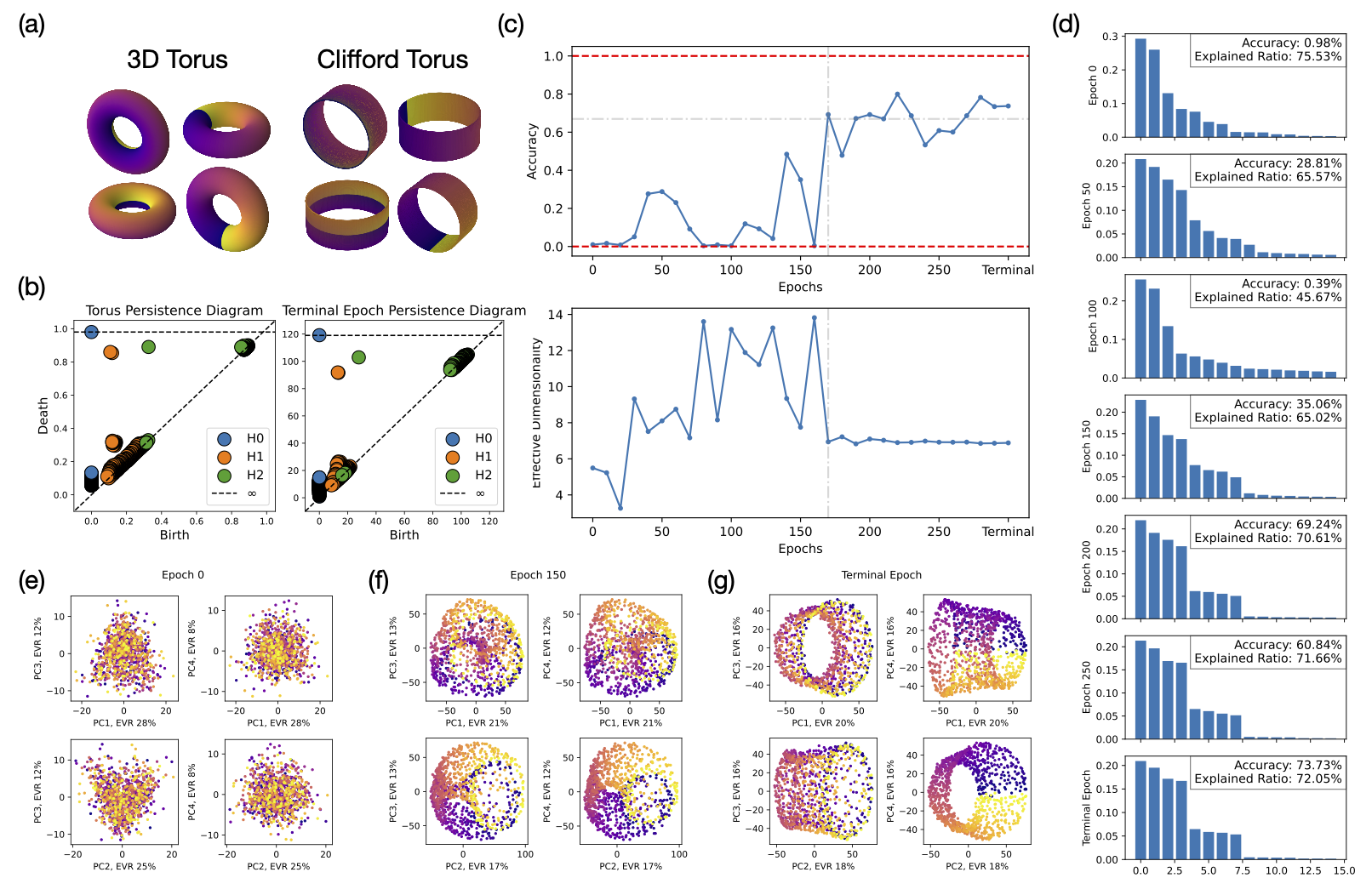}
\end{center}
\setlength{\belowcaptionskip}{-10pt}
\caption{\textbf{Metrics of a model trained using 2D Gaussian bump datasets with periodic boundaries.} \textbf{(a)} 2D projections of a standard 3D torus (left) and a 4D Clifford torus (right). The 3D torus is an example of a coupled representation that can be learned by the model while the 4D torus is a factorized one. \textbf{(b)} Persistence diagrams of a standard torus (left, the diagram looks the same for Clifford tori) and the learned representation of the model at the terminal epoch (right). There are two overlapping orange points for $H_1$ in both diagrams. \textbf{(c)} Model accuracy (top) and effective dimension (bottom) of representation learned by the model as a function of training epochs. \textbf{(d)} PCA eigenspectrum (the first 15 dimensions) of the model's learned representations and their corresponding sample accuracy percentage and explained variance ratio of the top 4 PCs (labeled top right of each panel) at various checkpoints during training. \textbf{(e)-(g)} PCA visualizations of the learned representations at epoch 0, 150, and terminal epoch, respectively.}
\label{fig:torus}
\end{figure}

In this section, we aim to understand the factorization of the model's learned representation via explicit inspection of its topology and geometry. Given a conditional DDPM trained on the 2D Gaussian \textbf{bump} dataset described above, we first investigate whether the model learns a coupled or factorized representation. Naïvely, a 2D Gaussian dataset with two independent features of variation, $x$ and $y$, has a 2D plane-like latent representation. Unfortunately, simply inspecting the learned representation would not allow us to easily differentiate between a coupled versus a factorized representation since the Cartesian product of two lines $L^1\times L^1 \subset \mathbb{R}\times \mathbb{R}=\mathbb{R}^2$ is topologically and geometrically equivalent to a plane $P^2\subset \mathbb{R}^2$. To overcome this issue, we perform a simple modification to the Gaussian dataset. We impose periodic boundary conditions in the image space to connect the left-right and top-bottom boundaries of the image such that the latent representation of the dataset forms a torus. Mathematically, a torus is defined by the Cartesian product of two circles $S^1\times S^1$. A Clifford Torus is an example of a factorized representation of the torus, with each circle independently embedded in $\mathbb {R}^2$, resulting in a 4-dimensional object. However, the most efficient representation in terms of extrinsic dimensionality is the regular torus $T^2$, which is embedded in $\mathbb{R}^3$, albeit unfactorized. Various 2D projections of the 3D torus and the Clifford (4D) torus are shown in Fig.~\ref{fig:torus}(a). Due to geometric differences between the Clifford and the 3D torus, we can now distinguish whether the model learns a factorized representation. If the model were to represent $x$ and $y$ independently with two ring manifolds, we expect to see a Clifford torus in the neural activations, rather than a different geometry such as the 3D torus.

To apply the geometry tests for differentiating between the 3D and the Clifford torus, we first need to confirm that the model indeed learns a torus representation of the dataset by computing the topological features of the learned representation via persistent homology. In Fig.~\ref{fig:torus}(b), we compare the persistence diagrams of a standard torus (left) with the final learned representation of the model (right). Both diagrams exhibit the similar topological features, rank-1 $H_0$ group, rank-2 $H_1$ group (with overlapping orange points on top in both diagrams), and rank-1 $H_2$ group, signaling that the model indeed learns a torus representation. 

We then investigate when this torus representation emerges during training. In Fig.~\ref{fig:torus}(c), we show the model's performance as measured by the accuracy of its generated images (top) and the corresponding effective dimension (bottom) of the learned representations across training, as measured by the participation ratio ( given by $(\sum_i \lambda_i)^2 / \sum_i \lambda_i^2$, where $\lambda_i$ is the eigenvalue of the $i$-th principal component \cite{Gao214262}). Intuitively, the participation ratio counts the dominant eigen-components, providing a measure of the effective dimension of the geometric object of interest. A detailed inspection of the effective dimension of the learned representations over the training duration informs us of whether and how the model arrives at a factorized, 4D representation. We see in Fig.~\ref{fig:torus}(c) that as training progresses, the model's internal representation first undergoes a dimension increase then a decrease, eventually converging to around 7 dimensions after 200 epochs. While the dimensionality converged to a higher dimension than 4, the top 4 eigenvectors became notably more prominent as the training converges, as indicated in the eigenspectra in Fig.~\ref{fig:torus}(d). This signals that a 4D, rather than 3D, torus representation is eventually learned. Various PCA projections of the learned representations along the training process are shown in Fig.~\ref{fig:torus}(g)-(i), where the projections in the last epoch resembles those of the Clifford torus shown in (a).

\begin{figure}[h]
\begin{center}
\includegraphics[width=5in]{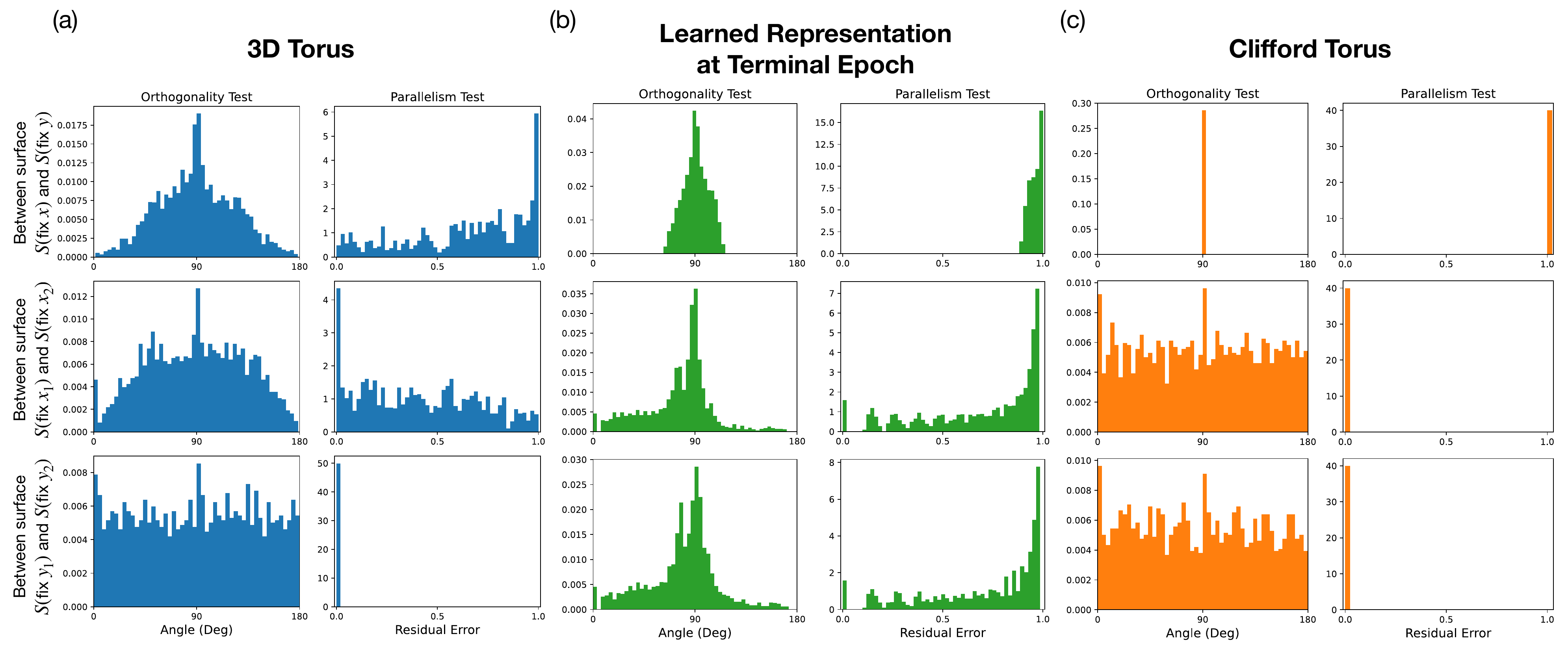}
\end{center}
\setlength{\belowcaptionskip}{-10pt}
\caption{\textbf{Comparison of orthogonality and parallelism test statistics between 3D torus, model's learned representation, and Clifford torus.} $x$-on-$y$ (top row), $x$-on-$x$ (middle row), $y$-on-$y$ (bottom row) orthogonality (left column) and parallelism (right column) test statistics are compared between (a) an ideal 3D torus (blue), (b) the learned representation by the model (green), and (c) an ideal Clifford torus (orange).}
\label{fig:torus_hist}
\end{figure}

To confirm whether the model learns a Clifford torus, we employ the orthogonality and parallelism tests of tori~\cite{torus_paper} (details found in Appendix~\ref{app-sec:torus}). In an ideal Clifford torus, the rings along the poloidal direction should be parallel with each other, and similarly for those along the toroidal direction. Moreover, the rings along the poloidal direction should be orthogonal with respect to the rings in the toroidal direction. The orthogonality and parallelism tests measure the orthogonality and parallelism of subspaces spanned by various rings on the torus. If the model learns a Clifford torus representation, the subspaces that code for $x$ and $y$ should be orthogonal, and the subspaces that code for $x$ ($y$) for any given $y$ ($x$) should be equivalent (parallel). 
Given a pair of $x$ and $y$, we construct 2D subspaces of rings $S(\text{fix } x)$ and $S(\text{fix } y)$ spanned by fixing $x$ while varying $y$ and fixing $y$ while varying $x$, respectively. We perform 3 sets of tests: 1) \textit{$x$-on-$y$ tests} test orthogonality and parallelism between surfaces $S(\text{fix } x)$ and $S(\text{fix } y)$ for all combinations of $x$ and $y$; 2) \textit{$x$-on-$x$ tests} test orthogonality and parallelism between $S(\text{fix } x_1)$ and $S(\text{fix } x_2)$ for all combinations of different $x$'s; 3) \textit{$y$-on-$y$ tests} test orthogonality and parallelism between $S(\text{fix } y_1)$ and $S(\text{fix } y_2)$ for all combinations of different $y$'s. 

In Fig.~\ref{fig:torus_hist}, we compare the orthogonality and parallelism test statistics between an ideal 3D torus (left), the model's learned representation at the final epoch (center), and an ideal Clifford torus (right). Comparing test 1 statistics (top row) between the model representation and the standard 3D and Clifford tori, we see that the model encodes $x$ and $y$ independently in mostly orthogonal subspaces, similar to a Clifford torus. The test 2 and 3 statistics (middle and bottom row) of the model representation, however, does not match those of either the 3D or the Clifford torus. In fact, the statistics from tests 2 and 3 of the model representation closely resemble those from test 1 of the 3D torus. This means that the model is encoding different values of $x$ in an almost orthogonal, as opposed to parallel, fashion, similarly for different values of $y$. These observations signify that the models ``hyper-factorizes'' each individual factor of variation by treating them more similarly to categorical variables with nonzero overlaps between neighboring categories rather than fully-continuous variables, which explains why the effective dimension of the model's learned representation is higher than 4 after convergence in Fig.~\ref{fig:torus}(c). Although these observations are made from a model trained on a torus dataset, we expect the model to have similar encoding schemes across all variants of our Gaussian bump datasets. 

\subsection{Models can compose but not interpolate}
\label{sec:res-composition}

\begin{figure}[!htb]
\begin{center}
\includegraphics[width=5in]{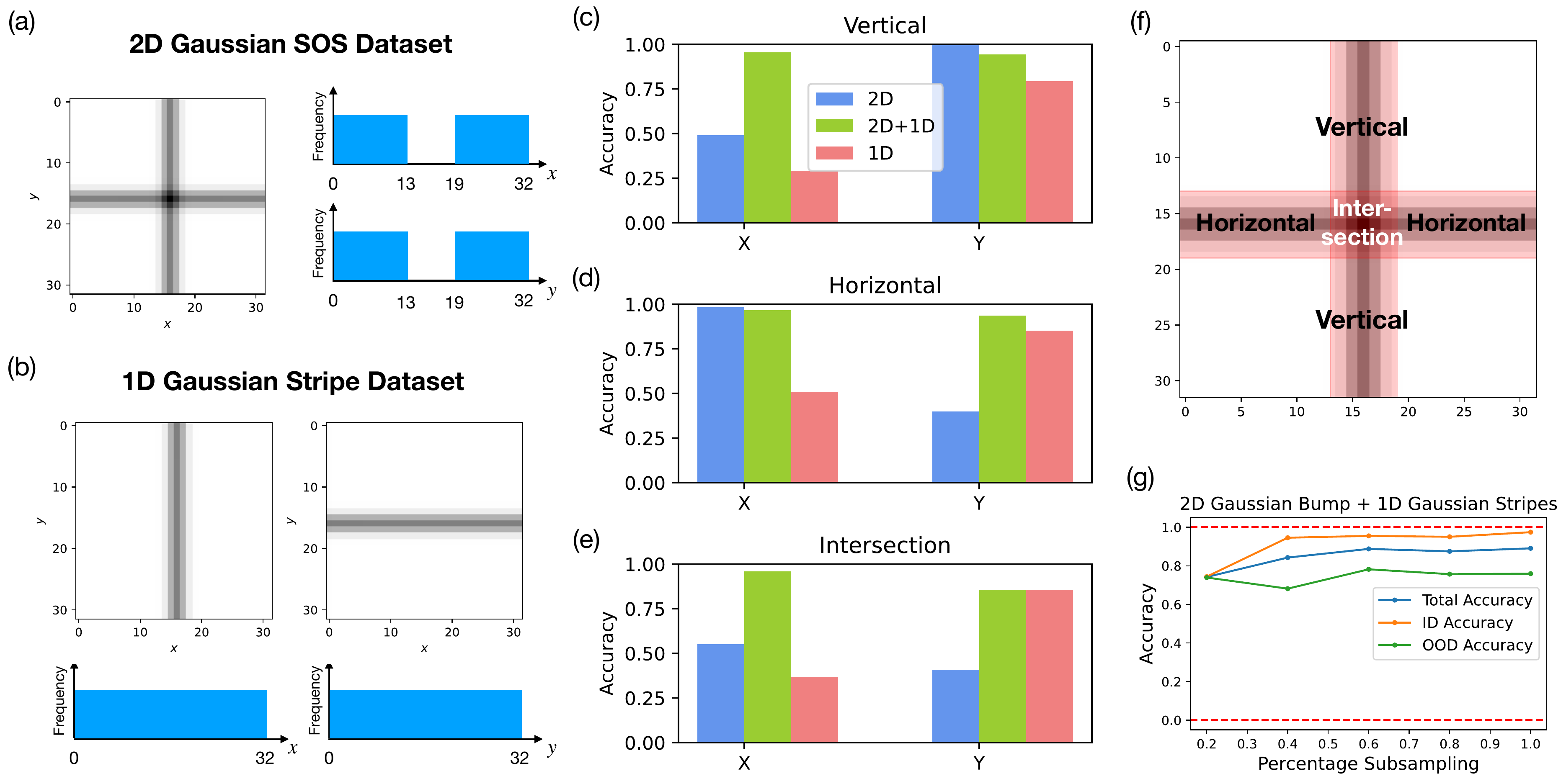}
\end{center}
\setlength{\belowcaptionskip}{-10pt}
\caption{\textbf{Models trained on Gaussian SOS datasets to generalize to the test regions.} We train three models on various Gaussian SOS datasets to test their ability to compositionally generalize in the red-shaded, held-out test regions shown in the sample image (f). \textbf{(a)} The 2D Gaussian \textbf{SOS} dataset contains all combination of 2D Gaussian SOSs for all $x$ and $y$ between 0 and 32 except for the held-out range between 13 and 19. \textbf{(b)} The 1D Gaussian \textbf{stripe} dataset contains horizontal and vertical 1D Gaussian stripes of full range of $x$ and $y$ values between 0 and 32. The accuracy of the three models in generating the correct $x$ and $y$ location of the Gaussian SOS is shown for different sections of the test regions: \textbf{(c)} The vertical section excluding the intersection, \textbf{(d)} the horizontal section excluding the intersection, and \textbf{(e)} the intersection. \textbf{(f)} Sample image of a 2D Gaussian SOS with the different test regions labeled. \textbf{(g)} shows the accuracy of models run with various subsampling rate of the 2D Gaussian bump + 1D Gaussian stripe dataset.}
\label{fig:flag}
\end{figure}

In this section, we examine the model's ability to compositionally generalize. Specifically, we train the model on incomplete datasets of 2D Gaussian SOSs, in which we leave out all Gaussian SOSs centered in the red-shaded test regions (Fig.~\ref{fig:flag}(f), sampling distributions shown in Fig. \ref{fig:flag}(a), (b)). We then assess the performance of the models in generating 2D Gaussian SOSs centered within and outside of the test regions. Here we choose the width of the cuts to be around 6 pixels wide, which roughly corresponds to the width of the Gaussian stripes of $\sigma=1.0$. We design the lesions such that we can probe the model's ability at compositionally generalizing out of the training distribution as well as its ability to spatially interpolate in a single variable alone. There are 4 possible outcomes based on the model's ability to compose and interpolate, and we give predictions for the model's generalization performance in each case: 1) the model cannot compose or interpolate: here, we would see low performance across the test regions; 2) the model interpolates but cannot compose: here, we would see high performance across the test regions; 3) the model composes but cannot interpolate: here, we would see high performance in one dimension but not the other in the non-intersecting part of the test regions, low performance in the intersection; 4) the model can compose and interpolate: here, we would see higher performance across the test regions. We note that case (2) and (4) are indistinguishable via the behavior of the model.

In our first experiment, we simply train our model on a 2D Gaussian SOS dataset where all data centered in the test regions are left out, i.e. excluding all $\mu_x, \mu_y\in[13,19]$ as shown in Fig.~\ref{fig:flag}(a). We call this model the \textit{2D model} since the model is trained on only 2D Gaussian SOS data. We then examine the terminal accuracy of the 2D model in generating Gaussian SOSs at the correct $\mu_x$ and $\mu_y$ in various parts of the test regions, which we section into a horizontal part, a vertical part, both excluding the area of their intersection, and the intersection itself (shown schematically in Fig.~\ref{fig:flag}(f)). We note that the 2D model achieves high accuracy in generating $\mu_y$ while suffers low accuracy in generating $\mu_x$ in the vertical section. Similarly, the 2D model achieves high accuracy in generating $\mu_x$ while suffers low accuracy in generating $\mu_y$ in the horizontal section. The model suffers low accuracy in generating both $\mu_x$ and $\mu_y$ in the intersection region. These observations have two implications: i) the model is factorized and compositional since it is able to generate the correct $\mu_x$ or $\mu_y$ irrespective of the other; ii) the model has limited ability to spatially interpolate, which suggests that it does not learn a fully continuous manifold in its activation space (see Fig.~\ref{fig:comparison}(c)). These observations meet our expectation for outcome case (3), when the model can compose but not interpolate and resonates our conclusion from Sec.~\ref{sec:res-factorization} that the model has learned to factorize \textit{x} and \textit{y}, but has not learned a consistent representation across all \textit{x}'s (and likewise \textit{y's}). In Appendix~\ref{sec:interpolation} Fig.~\ref{fig:interpolation}, we quantitatively assess model's ability to interpolate as a function of the held-out range width in the data manifold and found that model's ability to interpolate gradually decrease as a function of the held-out region width. 

To further verify our hypothesis that the model is capable of composition but not interpolation, we design an alternative dataset that includes a mixture of 2D Gaussian SOS data and 1D Gaussian stripe data (single stripes that are either horizontal or vertical). Specifically, we generate a set of 1D Gaussian stripe data across the entire range of $x$ and $y$ without any held-out regions, as shown in Fig.~\ref{fig:flag}(b). Note that the 1D Gaussian stripe data respects the 2D structure of the conditional input embedding. The exact generation procedure for the 1D Gaussian stripe dataset is described in Appendix~\ref{sec:stripe_data_gen}. This 1D dataset is then combined with the previous 2D Gaussian SOS dataset with the same held-out test regions. The resulting 2D+1D dataset contains 1D data in all range of $x$ and $y$ but has not seen 2D data (composition of 1D) in the test regions. The model trained under this dataset, deemed the \textit{2D+1D model}, has high performance in generating both $\mu_x$ and $\mu_y$ across all of the test regions. This shows that given the necessary ``ingredients" (1D Gaussian stripes of all range of $x$ and $y$), the model is able to compose them in a proper manner (2D Gaussian SOSs). These observations again confirm that the learned representation of the model is factorized but not consistent across a single feature, meaning that the model is able to compose independent features but not interpolate within a single feature dimension.

Finally, we test whether compositionality can be learned given no compositional examples in the training dataset. To test this, we train a \textit{1D model} on the 1D Gaussian stripe dataset as described above. Our results show that while model achieves respectable accuracy in generating the correct $\mu_y$ across the test regions, it does not simultaneously generate the correct $\mu_x$, indicating that the model did not learn to properly compose the 1D stripes. This is also evident from the generated image samples from the 1D model shown in Fig.~\ref{fig:comparison}(c). Without any compositional examples, the model fails to generate the intersection of two 1D Gaussian stripes and defaults back to generating one of the stripes at the correct location. While the 1D model has also learned to factorize $x$ and $y$ as independent concepts, it is not able to generalize here due to the lack of compositional examples. Hence, based on the performance of the 1D, 2D, and 2D+1D models that we have seen above, we conclude that complete compositional generalization requires data of all range of $x$'s and $y$'s and some compositional examples of the two to be present within the dataset.
\begin{figure}[!htb]
\begin{center}
\includegraphics[width=5in]{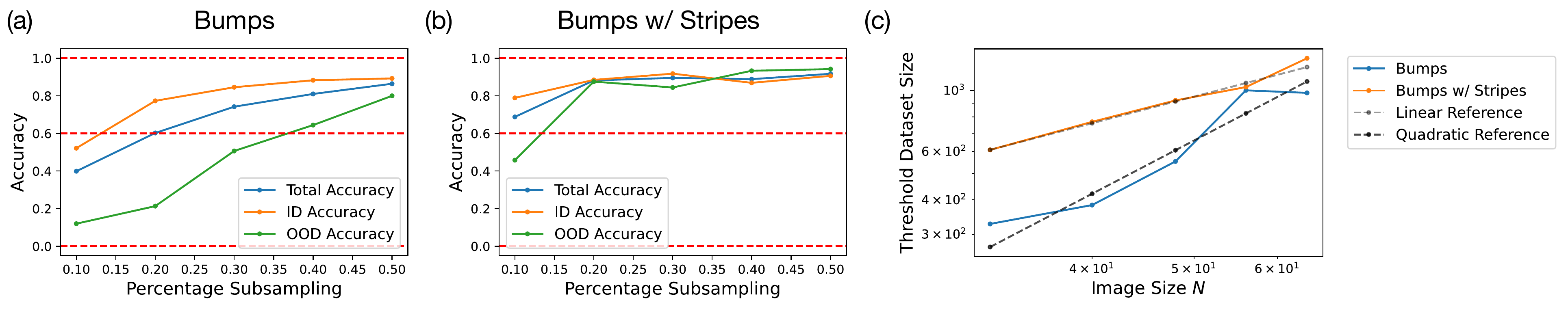}
\end{center}
\setlength{\belowcaptionskip}{-10pt}
\caption{\textbf{Sample efficiency gains from training the model on independent factors of variation.} \textbf{(a-b)}: Results on $N=32$ Gaussian 2D bump generation. \textbf{(a)} Model accuracy in generating 2D Gaussian bumps from training on 2D Gaussian bumps, shown as a function of the subsampling percentage. \textbf{(b)} Model accuracy in generating 2D Gaussian bumps from training on mixed 2D Gaussian bumps + 1D Gaussian stripes. Red dashed lines in (a),(b) mark a threshold accuracy of 0, 60, and 100\%. \textbf{(c)} Log-log plot of dataset size needed to reach 60\% threshold accuracy as a function of image size $N$ with 2D Gaussian bumps training data (blue) versus mixed 2D Gaussian bumps + 1D Gaussian stripe training data (orange): distinct scalings of data efficiency visualized by dashed gray and black lines, which provide a linear and quadratic reference, respectively. }
\label{fig:data_efficiency}
\end{figure}

Building on our previous results, we investigate whether combining the 2D Gaussian bump dataset with 1D stripe data can impart a different form of compositionality to the model -- multiplicative rather than additive. To do this, we create a dataset that mirrors the 2D+1D Gaussian SOS set, substituting the 2D Gaussian SOS data with 2D Gaussian bump data while maintaining the same test regions. The latent representation of this dataset can be found in Fig.~\ref{fig:data_latents}(h) in the Appendix. The resulting model trained under the 2D Gaussian bump + 1D Gaussian stripe dataset is able to generate 2D Gaussian bumps within the test regions up to 70\% accuracy, suggesting that the model is adept at handling an alternative form of compositionality. We further investigated whether the number of 2D Gaussian bump examples in the dataset influences compositionality by subsampling the 2D Gaussian bumps in the in-distribution regions (Fig. \ref{fig:data_latents}(g)). The results show that models trained on datasets with varying percentages of subsampled 2D Gaussian bumps achieve similar performance across the test regions, even with as little as 20\% of the original data. These findings suggest that training for compositional generalization can be efficiently achieved by using a full range of data for each individual factor of variation, along with a few compositional examples.

To validate the benefit of training with 2D + 1D datasets, we analyzed the data efficiency scaling of models trained on 2D Gaussian bumps + 1D Gaussian stripes, comparing them to models trained solely on 2D Gaussian bumps as a function of image size $N$. For both the baseline and augmented datasets, we excluded $20\%$ of the 2D Gaussian bumps from the center square region of the $N\times N$ canvas for out-of-distribution (OOD) evaluations. The accuracy of the models are plotted against the percentage of 2D Gaussian bumps subsampled in Fig.~\ref{fig:data_efficiency}(a) for the baseline dataset and (b) for the augmented dataset. The results clearly demonstrate that the model trained on the augmented dataset performs better with a significantly lower percentage of 2D Gaussian bumps in the training dataset. We then plotted the sizes of the baseline and augmented datasets required to reach a $60\%$ accuracy threshold as a function of $N$ on a log-log scale in Fig.~\ref{fig:data_efficiency}(c). We found that models trained on the augmented datasets only require linear number of data in $N$ while those trained on the baseline datasets require quadratic number of data in $N$. This result suggests that augmented datasets with samples of all independent factors plus a few compositional examples provide an efficient training method that has sample complexity linear in the number of factors. 
In Appendix~\ref{sec:comp} and~\ref{sec:composition}, we further discuss the sample efficiency of learning compositionality from dataset of isolated factors of variation plus few compositional examples. 

In summary, our findings indicate that the model excels at compositionality but struggles with interpolation. Consequently, effective generalization necessitates that the training set encompasses the full range of each individual factor of variation, alongside compositional examples. Furthermore, we discovered that models trained on datasets featuring explicitly factorized examples exhibit remarkable data efficiency and adaptability to various forms of compositionality. These results align with the observed factorization patterns in the model representation discussed in Sec.~\ref{sec:res-factorization} and suggest promising strategies for developing more data-efficient training schemes that utilize tailored datasets.

\subsection{Connection between manifold learning and percolation theory}
\label{sec:res-perc}
\begin{figure}[!htb]
\begin{center}
\includegraphics[width=5in]{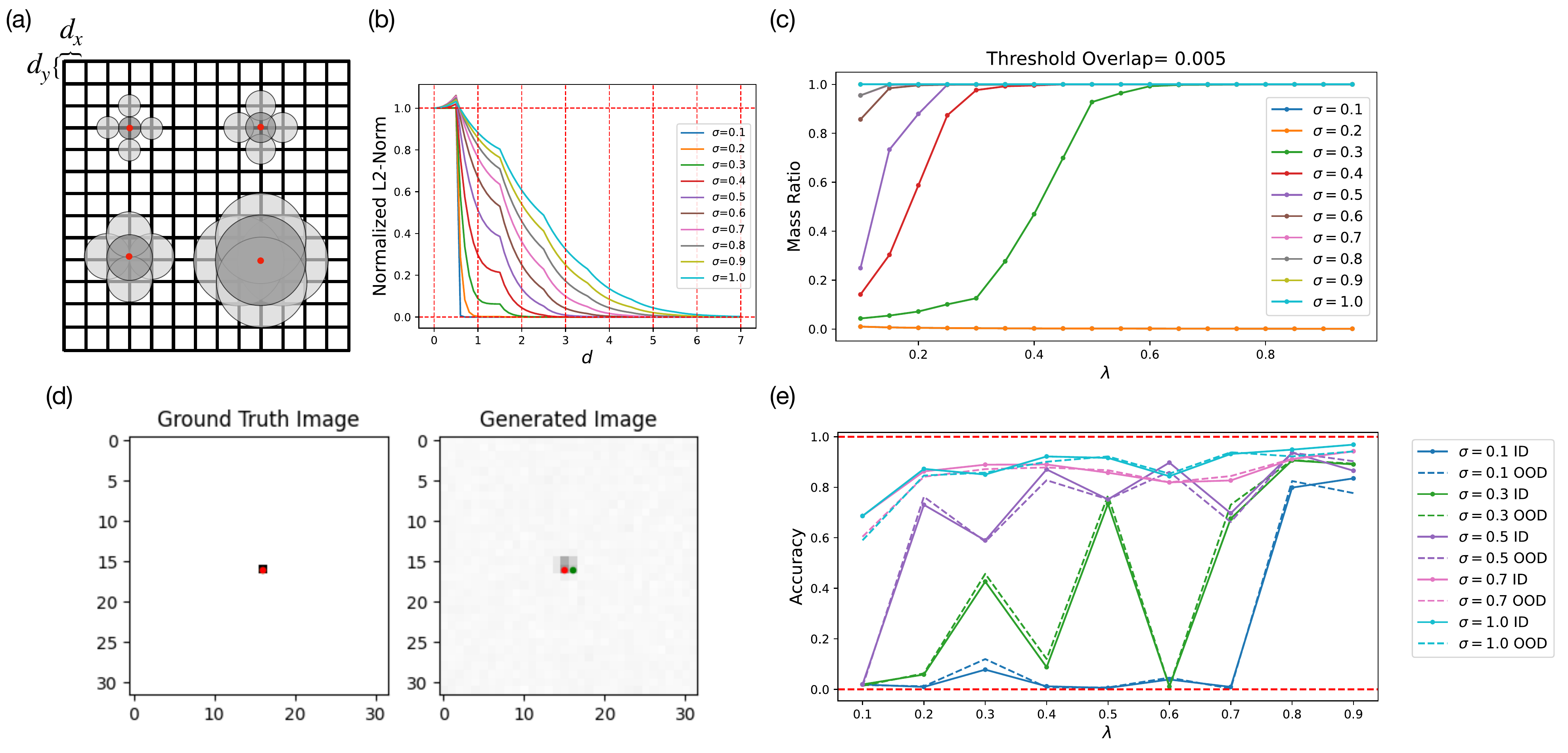}
\end{center}
\setlength{\belowcaptionskip}{-15pt}
\caption{\textbf{Percolation theory of manifold formation and interpolation.} \textbf{(a)} Schematic drawing of the Gaussian bumps of various width on a lattice of grid size $d_x$ and $d_y$. \textbf{(b)} Overlaps of neighboring Gaussian bumps as a function of the grid size $d:=d_x=d_y$ for various Gaussian widths. \textbf{(c)} Theoretical simulation of largest connected mass ratio as a function of percentage of data in training set $\lambda$ with threshold overlap of 0.005 for various Gaussian widths. \textbf{(d)} Ground truth (left) and generated image (right) of a sample Gaussian bump data of $\sigma=0.1$ centered at $\mu_x=16$ and $\mu_y=16$. \textbf{(e)} Terminal accuracy of models as a function of $\lambda$ trained with datasets that only differ in the Gaussian bump widths for various Gaussian bump widths.}
\label{fig:percolation}
\end{figure}

We have established that the model learns factorized (but not fully continuous) and compositional representations. Next, we identify a potential connection between manifold learning in diffusion models and percolation theory in physics, which provides a normative explanation as to how factorization (or compositional generalization) emerges. To start, we note that our Gaussian bump datasets can be approximated as a simple Poisson Boolean (Gilbert's disk) model \cite{gilbert_disk, boolean_poisson_model}, a well-studied system in continuum percolation theory. Fig.~\ref{fig:percolation}(a) shows a schematic illustration of our dataset of Gaussian bumps approximated as disks of various widths on a 2D $32\times 32$ lattice of grid spacings $d_x$ and $d_y$. In percolation theory, the quantity of interest is the critical fraction of nodes that need to have non-zero overlaps in order for the entire system to be interconnected with high probability. For most systems (finite- and infinite-sized), there exist a phase transition that occurs at the critical fraction that can be either analytically derived or numerically estimated, with the transition becoming sharper as the system size scales. In our context, since the Gaussian bump images are pixelated, overlaps between neighboring Gaussian bumps are not smooth. In Fig.~\ref{fig:percolation}(b), we plot the normalized L2-norm of the neighboring distance-$d$-separated Gaussian bumps with various spread $\sigma$'s. We note that since the L2-norm measure of overlap is normalized to 1 between a given Gaussian bump and itself, the overlap between a given Gaussian bump and a slightly offset Gaussian bump temporarily goes beyond 1 due to the discrete nature of the data. 

We hypothesize that below a threshold amount of training data, the diffusion model cannot construct a faithful representation of the training dataset. From the percolation perspective, if there does not exist a large enough interconnected component within the dataset, the model will fail to learn the relative spatial location of the data points, making it hard to learn a faithful 2D representation. To test this hypothesis, we first simulated the mass ratio of largest interconnected components as a function of the unit area data density $\lambda$ with our 2D Gaussian bump datasets. For simulation, we use a dataset of 1024 data points, corresponding to $d:=d_x=d_y=1.0$ on a $32\times 32$ lattice. We then randomly sample $\lambda\times 1024$ points on the lattice to compute the size of the largest interconnected cluster of Gaussian bumps. Here, we define a hyperparameter of threshold overlap beyond which we consider two Gaussian bumps as overlapping. In Fig.~\ref{fig:percolation}(c), we show the simulation results. averaged over 5 runs, with the chosen threshold overlap of 0.005 for datasets with various $\sigma$'s. Additional simulation results with various chosen overlap thresholds can be found in Appendix~\ref{app-sec:percolation}.  

Next, we quantify the connection between percolation theory and manifold formation in diffusion models. To account for the stochasticity in sampling the datasets and avoid significant overhead in model training, we choose a fix set of grid points on the $32\times 32$ lattice and generate datasets of varying $\sigma$'s from this fixed set of grid points. Moreover, we choose each sampled fraction of the dataset to be a strict superset of the smaller sampled fractions of the dataset. For example, dataset at $\lambda=0.2$ contains all of dataset at $\lambda=0.1$, and dataset at $\lambda=0.3$ contains all of dataset at $\lambda=0.2$ and hence all of $\lambda=0.1$, so on and so forth. For datasets of different $\sigma$'s, we chose the same grid points such that the only difference between datasets of different $\sigma$'s is just the $\sigma$ itself. This procedure eliminates all the stochasticity due to the probabilistic nature of percolation, and ensures that differences in training are only due to the differences in $\sigma$ and $\lambda$. To ensure fair comparison, we proportionally add training time to each model trained with fewer data such that the total amount of training batches are constant across all models. 

In Fig.~\ref{fig:percolation}(e), we display the accuracy of models trained with different fractions of a training dataset of a total size of 1024 for various $\sigma$'s. We note that there are high variances in the model performance, especially for $\sigma=0.3, 0.5$. This is due to stochasticity in model training itself as well as the gradual as opposed to a sharp phase transition, as shown in simulation Fig.~\ref{fig:percolation}(c). In addition, we notice that based on the simulation, $\sigma=0.1$ will not forge any connected cluster due to its minuscule spread. In Fig.~\ref{fig:percolation}(d), an sample image of Gaussian bumps of $\sigma=0.1$ is shown on the left, where it is a mere pixel centered at $(\mu_x, \mu_y)$. Nonetheless, in experiment Fig.~\ref{fig:percolation}(e), we see that the model is able to learn the $\sigma=0.1$ dataset at $\lambda=0.8, 0.9$. This is partially due to the fact that the noising process in diffusion training smears out the Gaussian bump of $\sigma=0.1$ to give it a larger effective width, as shown in the generated bump image in Fig.~\ref{fig:percolation}(d). In general, our experimental results agree with the percolation theory prediction that smaller similarities between the samples makes it harder for the models to learn meaningful representations, despite the same quanitity of data. The portraits of the learned representations by the models generating Fig.~\ref{fig:percolation}(e) is shown in Fig.~\ref{fig:perc_manifolds} of Appendix~\ref{sec:percolation_exp}. 

\section{Discussion}
\label{sec:discussion}
In the previous section, we have shown that diffusion models are capable of learning factorized and semi-continuous representations. This allows for compositional generalization across factors that have appeared in the training dataset, even given only a small number of compositional examples. While we studied a toy system, our results imply that the diffusion model architecture has some inductive biases favoring factorization and compositionality, as seen in astonishing compositional text-to-image generation examples such as an ``astronaut riding a horse on the moon". Our results demonstrate further that if the training data include isolated factors of independent variation and some compositional examples, diffusion models are capable of attaining high performance in OOD compositional generalization. While it is true that natural images are much more complex, and there can be numerous forms of compositionality within a single image, the datasets and modes of composition we studied were not trivial. Nonetheless, further investigation is necessary to understand the compositionality and factorization of models when multiple forms of compositionality are at play. 

Furthermore, we showed a connection between the models' performance, as related to its ability to learn a faithful representation of the dataset, and percolation theory in physics. Percolation theory provides a plausible mechanistic interpretation of the observed sudden emergence in the model's capability beyond a threshold number of data points. Further work is needed to characterize the precise connection between percolation theory and manifold formation, in both toy settings and realistic settings. In realistic datasets, mutual information or cosine similarity between data points can serve as abstract forms of overlap. Moreover, the idea of percolation can be further extended to study alternative observations of phase transitions in deep learning, such as percolation of the chain-of-knowledge in large language models.

\section{Conclusion}
\label{sec:conclusion}
We have shown that diffusion models are capable of learning factorized representations that can compositionally generalize OOD, given data containing the full range of each independent factor of variation and a small amount of compositional examples. Our study suggests that diffusion models have the inductive bias for factorization and compositionality, which are believed to be key ingredients for scalability. We identified that the diffusion models fail to generalize out-of-distribution when 1) there are unseen values of a given factor of variation for the composition, 2) there are no compositional examples in the training dataset, 3) there is an insufficient quantity of data, and 4) there is an insufficient amount of overlaps between data. Together, our results imply that a more efficient training dataset can be constructed by incorporating samples that feature explicit factorized examples along with a few compositional examples that have substantial overlap. By optimizing the training data in this manner, we can enhance the data efficiency scaling of diffusion-based models. Future investigations should consider controlled experiments with data optimization for improving data efficiency of diffusion models using more realistic data. 

\bibliography{bib}
\bibliographystyle{plainnat}


\appendix

\section{Experimental Details}
\label{app-sec:exp_details}
\subsection{Architecture}
\label{sec:app_architecture}
\begin{figure}[h]
\begin{center}
\includegraphics[width=5.0in]{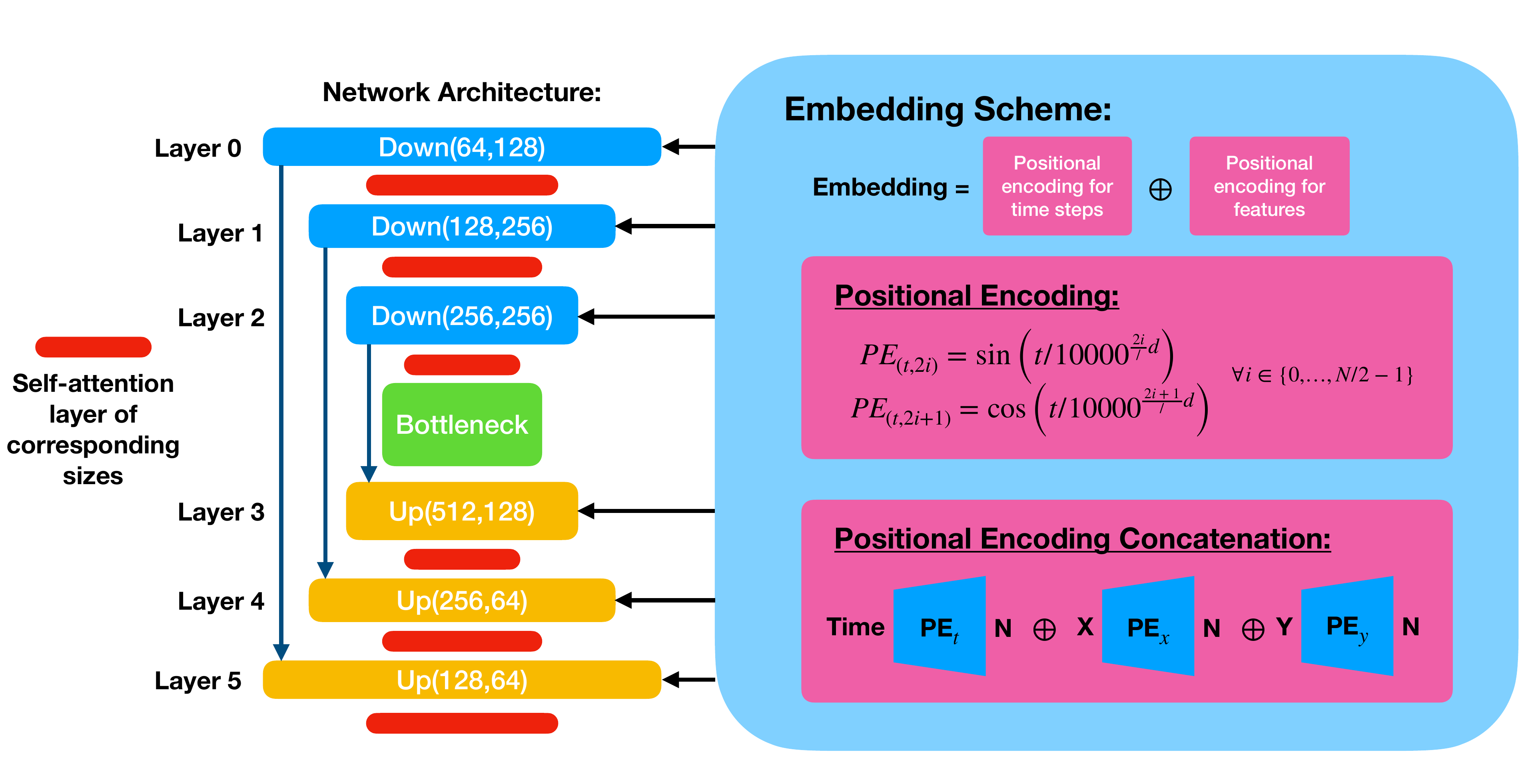}
\end{center}
\caption{\textbf{The UNet architecture of the conditional diffusion model.} The schematic diagram of the standard UNet architecture consisting of three downsampling/upsampling blocks with interlaced self-attention layers and skip connections is shown on the left. The conditional information consisting of a concatenation of positional encodings of timestep and $x$/$y$-positions is passed in at each block as shown on the right.}
\label{fig:architecture}
\end{figure}

We train a conditional denoising diffusion probabilistic model (DDPM)~\cite{ho2020denoising} with a standard UNet architecture of 3 downsampling and upsampling blocks, interlaced self-attention layers, and skip connections as shown in Fig.~\ref{fig:architecture}. Each down/up-sampling blocks consist of max pooling/upsampling layers followed by two double convolutional layers made up by convolutional layers, group normalization, and GELU activation functions. 

The conditional information is passed in at each down/up-sampling block as shown in the schematic drawing. In our experiments, we choose to preserve the continuity of the Gaussian position labels passed into the model via explicit positional encoding rather than using a separate trainable embedding MLP at each block. Each embedding vector is made by concatenating equal-length vectors of the positional encodings of the timestep, the $x$-position, and the $y$-position. 

In our experiments, we visualize the outputs of layer 4 as the internal representation of the diffusion model. We have chosen not to use the output of the bottleneck layer for our study of the learned latent manifold, as we have observed that the bottleneck layers have diminishing signals in most of our experiments. This is likely due to the presence of the skip connections. This choice does not affect the validity of our main results, as we are focused on the factorization of the model's learned representation.

\subsection{Dimension Reduction}
\label{sec:app_umap}
We primarily use the dimension reduction technique Uniform Manifold Approximation and Projection for Dimension Reduction (UMAP)~\cite{mcinnes2020umap} to study and visualize the learned representation of the model. Specifically, we collect image samples and their corresponding internal representations (outputs of layer 4 from the architecture described in Sec.~\ref{sec:app_architecture}). We then transform the high-dimensional internal representations into a 3D embedding as a sample of the learned representation, which we visualize and analyze. For an implementation of UMAP, we used the Python package~\cite{mcinnes2018umap-software}.

\subsection{Evaluation}
\label{sec:app_metrics}
We assess the performance of the model using two primary criteria: 1) the quality of the denoised images and 2) the quality of the learned representation. 

At a given time during or after training, we generate 1024 denoised images and their corresponding internal representations of sampled labels based on $32\times 32$ grid points. We predict the label corresponding to each generated image based on the $x$- and $y$-positions of the generated Gaussian bump/SOS in the image. We then compute the accuracy of predicted labels from the ground-truth labels averaged over 1024 samples as 
\begin{align}
    \text{Accuracy} = \frac{1}{1024}\sum_{i=1}^{1024} \boldsymbol {1}(|\mu^i_x-\hat{\mu}^i_x|<1)\cdot \boldsymbol {1}(|\mu^i_y-\hat{\mu}^i_y|<1),
\end{align}
where $\boldsymbol {1}(\cdot)$ is an indicator function that returns 1 if the expression within holds true, 0 otherwise. Similarly, we can modify this expression to only assess the accuracy of generated $x$-positions or $y$-positions separately. Here we estimate the center of the Gaussian bump/SOS $\hat{\mu}^i_x$ and $\hat{\mu}^i_y$ by finding the location of the darkest pixel in the image. In the cases where there are no Gaussian bumps/SOS's or more than one bump/SOS, the algorithm defaults back to finding the centroid of the image. We then construct the learned representation of the model based on the neural activations of layer 4 corresponding to the 1024 sampled images collected at the terminal diffusion generation timestep. We note that we have investigated neural activations collected at alternative diffusion generation timesteps and found that they do not noticeably differ from timestep to timestep.


\subsection{Training Loss}
Diffusion models iteratively denoise a Gaussian noisy image $\mathbf{x}_T$ into a noisefree image $\mathbf{x}_0$ over diffusion timesteps $t\in\{0,1,\ldots, T\}$ given the forward distribution $q(\mathbf{x}_t|\mathbf{x}_{t-1})$ by learning the reverse distribution $p_{\theta}(\mathbf{x}_{t-1}|\mathbf{x}_t)$. Given a conditional cue $\mathbf{c}$, a conditional diffusion model~\citep{cond1,cond2} reconstructs an image from a source distribution $q(\mathbf{x}_0|\mathbf{c})$. Specifically, we train our neural network (UNet) to predict the denoising direction $\epsilon_{\theta}(\mathbf{x}_t, t, \mathbf{c})$ at a given timestep $t$ with conditional cue $\mathbf{c}$ with the goal of minimizing the mean squared loss (MSE) between the predicted and the ground truth noise $\epsilon$ as follows
\begin{align}
    \mathcal{L}:=\mathbb{E}_{t\in\{0,\ldots, T\}, \mathbf{x}_0\sim q(x_0|\mathbf{c}), \epsilon\sim\mathcal{N}(\mathbf{0}, \mathbf{I})}\left[\lVert \epsilon - \epsilon_{\theta}(\mathbf{x}_0, t, \mathbf{c})\rVert^2\right],
\end{align}
where we assume each noise vector $\epsilon$ to be sampled from a Gaussian distribution $\mathcal{N}(\mathbf{0}, \mathbf{I})$ I.I.D. at each timestep $t$.

\subsection{Datasets}
\label{sec:app_dataset}

The datasets we used for training the models generating the results in Sec.~\ref{sec:results} have various increments $d$ and $\sigma$. Here we briefly comment on the interplay between increments and sigmas, and how they affect dataset densities and overlaps. The ultimate goal of our task of interest is to learn a continuous 2D manifold of all possible locations of the Gaussian bumps/SOS's. Intuitively, the spatial information necessary for an organized, continuous, and semantically meaningful representation to emerge is encoded in the overlap of the neighboring Gaussian bumps/SOS's, which is tuned via the parameters $d$ and $\sigma$. As we increase $d$, the size of the dataset gets scaled quadratically, resulting in denser tilings of the Gaussian bumps/SOS's. 

As we scale up $\sigma$, the dataset size remains fixed while the overlaps with neighbors are significantly increased. In Fig.~\ref{fig:percolation}(b), we plot the normalized L2-norm of the product image of neighboring Gaussian bumps as a function of increments for various spreads. Specifically, given two inverted grayscale Gaussian bump images, $a$ and $b$, the normalized L2-norm of their product is given by the formula $\lVert\sqrt{a * b}\rVert_2/\lVert a\rVert_2$, where $*$ is element-wise multiplication and $\lVert\cdot\rVert_2$ is the L2-norm. This quantity should give a rough measure of the image overlap with the exception at increment around 0.5 due to the discrete nature of our data. Moreover, we note that the cusps in the curves occur for the same reason. As we can see, the number of neighbors that a given Gaussian bump has non-trivial overlaps with grows roughly linearly to sub-linearly with the spread. Since the model constructs non-parallel representation encodings of different values of $x$'s and $y$'s, more neighbor information results in more overlaps between different values of $x$'s and $y$'s and hence better representation learned, as shown in the percolation results in Fig.~\ref{fig:percolation} in Sec.~\ref{sec:res-perc}. Overlaps in Gaussian SOS data can be similarly analyzed, albeit the differences in overlaps of neighboring Gaussian SOS's of various spread $\sigma$'s is not as pronounced due to the large coverage area of each Gaussian stripe.

\subsection{Training Details}
\label{sec:app_training}
We train the models on a quad-core Nvidia A100 GPU, and an average training session lasts around 6 hours (including intermediate sampling time). Each model is given an inversely proportional amount of training time as a function of the size of the dataset that it is trained on such that there is sufficient amount of training for the models to converge. For each model, we sample 1024 samples corresponding to the 32$\times$32 grid points of $\mu_x$ and $\mu_y$ pairs tiling the entire image space. The model is trained with the AdamW optimizer with built-in weight decay, and we employ a learning rate scheduler from Pytorch during training. We did not perform any hyperparameter tuning, although we would expect our observations to hold regardless. 

\section{Orthogonality and Parallel Test}
\label{app-sec:torus}
\begin{figure}[h]
\begin{center}
\includegraphics[width=5.8in]{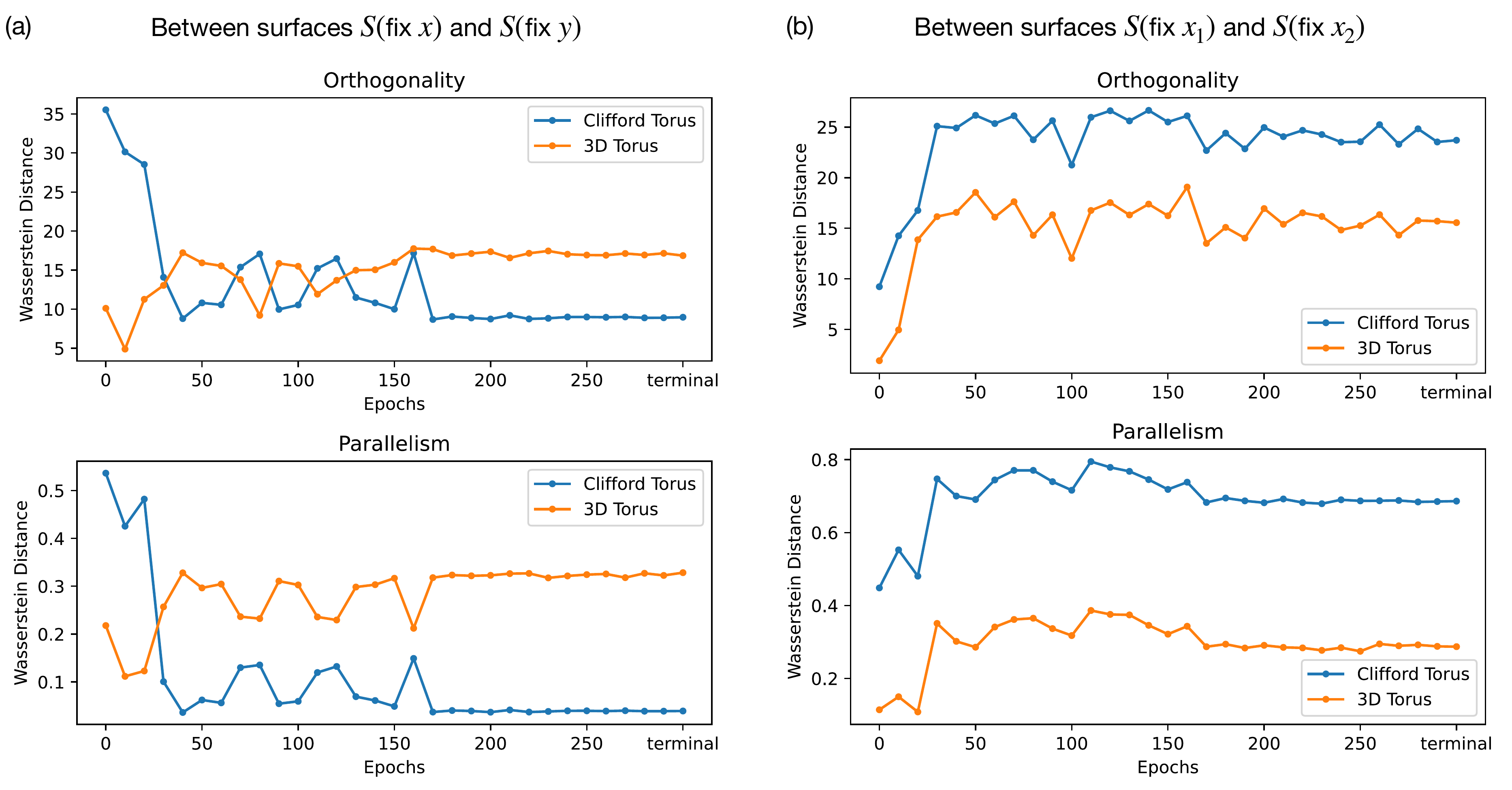}
\end{center}
\caption{\textbf{Wasserstein distances of orthogonality and parallelism tests statistics between learned representations and ideal Clifford/3D tori. } }
\label{fig:torus_wassertein}
\end{figure}

\subsection{Motivation for the Analysis}

To better identify whether the diffusion models learn to factorize its representations, we choose to study periodic representations of Gaussian bumps, which we expect to form toroidal manifolds in the latent space as opposed to planar manifolds. This is because toroidal manifolds have multiple standard realizations which vary based on their degree of entanglement in the two rings. In particular, the standard torus embeds the rings in three dimensions that couple the two periodic variables, whereas the Clifford Torus embeds each ring in its own orthogonal subspace. This difference in geometry motivates the two analyses that we implemented to identify which geometry the diffusion model learned. 

\subsection{Computation of the Tests}

We implement the orthogonality and parallelism tests as in \cite{cueva2021recurrent}. Briefly, the orthogonality test asks whether the rings that code for each variable lie in orthogonal subspaces--if so, this indicates that the torus has a Clifford-like geometry. The parallelism test asks whether rings in one variable are aligned in the other variable. If so, this also indicates that the torus has a Clifford-like geometry. 

The analysis first requires identifying each ring--surprisingly, we found that when the rings formed, the top four Principal Components comprised them. In particular, the first ring were consistently spanned by PCs 1 and 3, and the second was spanned by PCs 2 and 4 (Fig.~\ref{fig:torus}(f) and (g)). Thus, we confirmed that we could use PCA for our analysis. 

For each analysis, we identify the ring subspaces by fixing one variable of the input, and sweeping the other variable. We do so for each index, arriving at 32 different subsets of the data for variable $x$ and 32 different points for variable $y$. Next, we compute the top two Principal Components for each subset of the data, thereby identifying the ring. 

For the orthogonality test, we randomly draw pairs of rings that are fixed in $x$ and $y$ and compute the pairwise angles between each rings' Principal Components via cosine similarity. For the parallelism test, we compute the randomly drawn pairs of rings that are fixed in the same variable. Next, we create a projection matrix $P$ from the subspaces defined by one of the rings, and calculate the reconstruction error of the second ring's projection onto that subspace:
\begin{equation}
r = |h - PP^Th|^2_2.
\end{equation}

In each test, we aggregate histograms of the respective metric. To identify the similarity between our data and either toroidal embedding, we compute the same histograms for both the Clifford and standard torus (based on a simulation with similar spacing in $x$ and $y$ samples), and compute the Wasserstein distance between the data and these simulations. The Wasserstein distances between the orthogonality and parallelism test statistics of the model's learned representation and those of an ideal Clifford torus and 3D torus are shown in Fig.~\ref{fig:torus_wassertein} as a function of the training epochs. We note that the model's learned representations at smaller training epochs are not necessarily a torus, hence the metric comparison at those points has less significance.

\section{Composition and Interpolation}
\label{app-sec:interpolation}
\subsection{Gaussian Stripe/SOS Dataset Generation}
\label{sec:stripe_data_gen}

\begin{figure}[!htb]
\begin{center}
\includegraphics[width=5.5in]{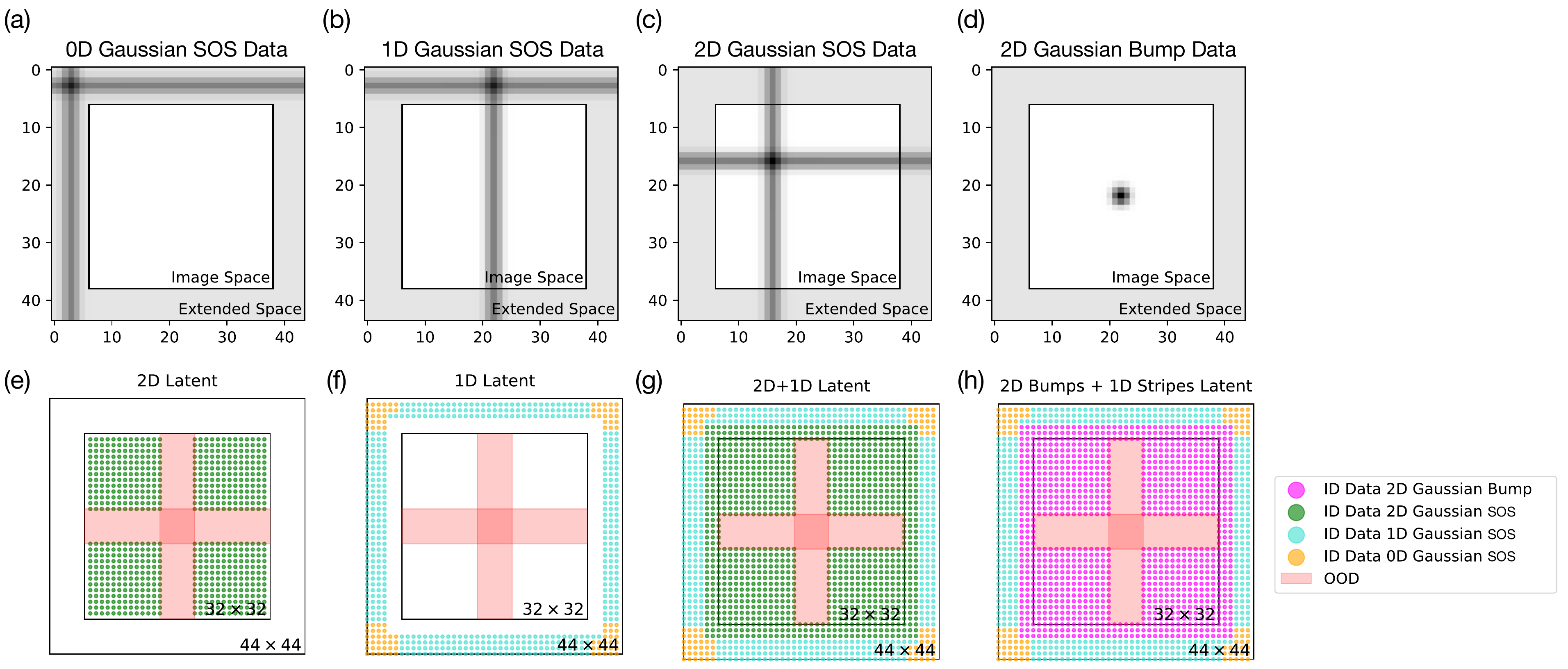}
\end{center}
\setlength{\belowcaptionskip}{-10pt}
\caption{\textbf{Sample Gaussian stripe/SOS/bump data images and dataset latent representations used for generating models in Fig.~\ref{fig:flag}.} We generate the 1D Gaussian stripes via embedding the $32\times 32$ image frame into an extended latent space of $44\times 44$, where the extended portion is not visible to the model. \textbf{(a)-(c)} show example images of 0D, 1D, and 2D Gaussian SOS data images, where the dimensionality of the image is defined by the dimension of the portion of data visible within the $32\times 32$ image frame. \textbf{(d)} shows the equivalent augmentation of 2D Gaussian bump data in the embedded space. For each data image, we assign its position labels from range 0 to 44 while keeping the image size to be $32\times 32$ by cropping out the ``extended space" in the image. Based on these image data, we construct the 2D, 1D, and 2D+1D datasets of Gaussian SOS's that we use to train the models in Fig.~\ref{fig:flag}. \textbf{(e)-(g)} show the latent representation of the 2D, 1D, and 2D+1D datasets respectively, where the green dots represent 2D Gaussian SOS data, the blue dots represent the 1D Gaussian SOS data, and the yellow points the 0D Gaussian SOS data. We leave all 2D Gaussian SOS centered in the red-shaded region out of the training dataset for our compositional generalization evaluation. \textbf{(h)} shows the equivalent dataset to the 2D+1D dataset with the 2D Gaussian SOS's replace with 2D Gaussian bumps, where the magenta points represent the 2D Gaussian bump data. }
\label{fig:data_latents}
\end{figure}

To generate 1D Gaussian stripe dataset shown in Fig.~\ref{fig:flag}(b) while maintaining the structure of the 2D conditional embeddings, we embed the $32\times 32$ data latent space into an extended $44\times 44$ latent space while maintaining the image pixel size to be $32\times 32$. Functionally, this allows us to mix into the 2D Gaussian SOS datasets 1D (and 0D) Gaussian data, as defined by the portion of the Gaussian that is actually visible to the model in the $32\times 32$ pixels. Examples of 0D, 1D, and 2D Gaussian SOS data are shown in Fig.~\ref{fig:data_latents}(a)-(c), respectively. Fig.~\ref{fig:data_latents}(e)-(g) then show the actual data latent representations of the 2D, 1D, and 2D+1D datasets that we use to train 2D, 1D, and 2D+1D models described in Sec.~\ref{sec:res-composition}. Here, the green points represent each individual 2D Gaussian SOS data, the blue points represent the 1D Gaussian SOS data, and the yellow points the 0D Gaussian SOS data. In all three datasets, we leave out the all Gaussian SOS data centered in the red shaded regions. We note that due to the nature of our data generation scheme, the 2D data only exist within the image space (and some spillover) and the 1D (and 0D) data only exist in the extended data latent space. Because of the nonzero width of the Gaussian stripes, a portion of the data along the rim of the $32\times 32$ image space is actually 2D with a portion of the overlaps between the horizontal and the vertical stripes visible within the image frame. In training the 1D model shown in Fig.~\ref{fig:flag}, we have removed the rim where any overlap between the horizontal and vertical Gaussian stripes is partially visible within the image, as shown in Fig.~\ref{fig:data_latents}(f). The model trained on such a 1D dataset without any compositional example is, as discussed in Sec.~\ref{sec:res-composition}, not able to handle the conjunction of the 2 Gaussian stripes, as shown in Fig.~\ref{fig:comparison}(a)-(c).

\subsection{Generalization with Few Compositional Examples}
\label{sec:comp}

\begin{figure}[h]
\begin{center}
\includegraphics[width=5.4in]{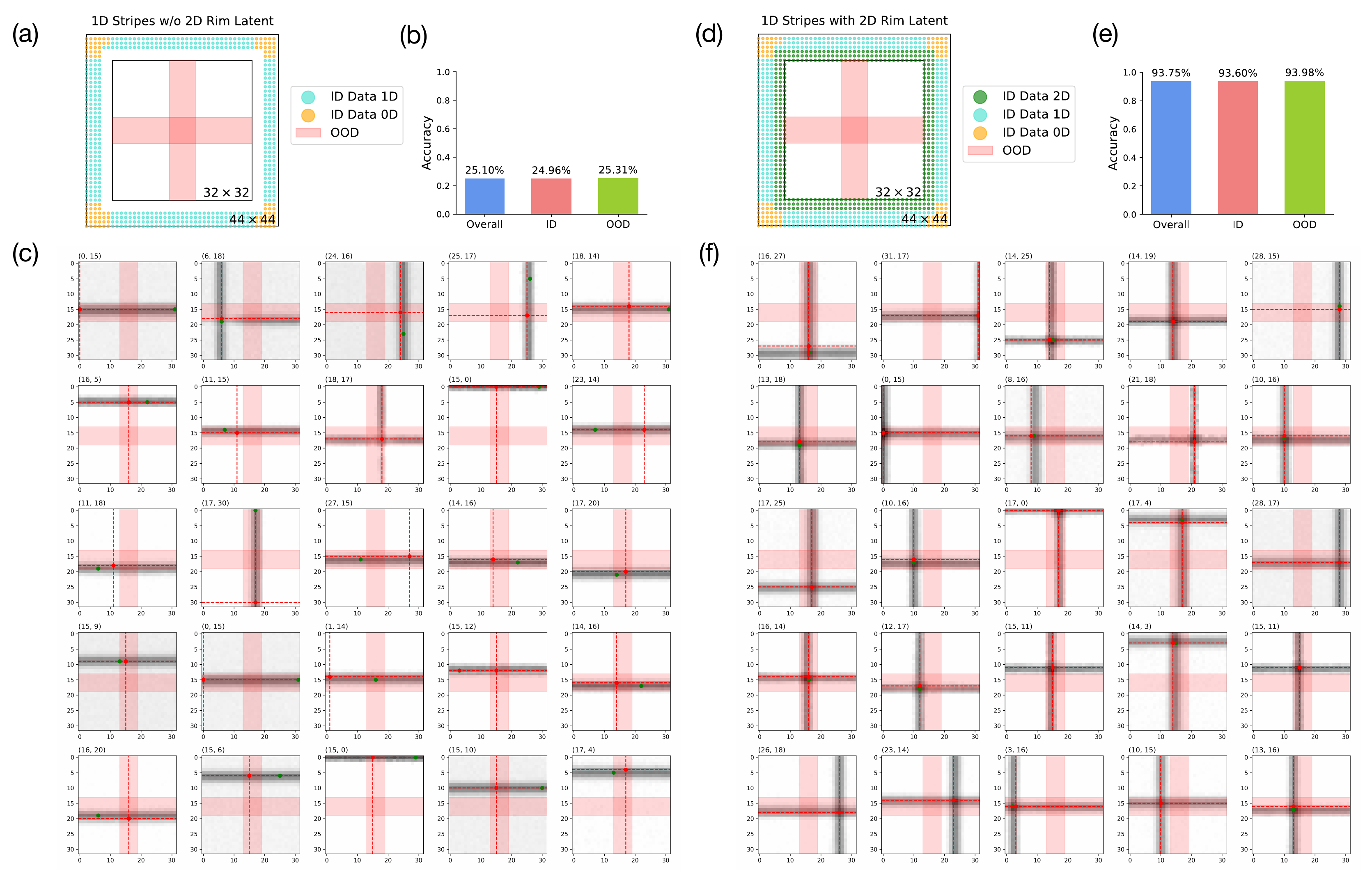}
\end{center}
\caption{\textbf{Comparison between 1D Gaussian stripe dataset with and without 2D Gaussian SOS examples.} \textbf{(a)} shows the latent manifold of a pure 1D Gaussian stripe dataset with the rim of 2D Gaussian SOS examples removed along the $32\times 32$ image space. The overall, ID, and OOD accuracy of the generate images of the model trained under this dataset is shown in \textbf{(b)} along with 25 randomly selected generated images in \textbf{(c)}. \textbf{(d)} shows the latent manifold of the 1D Gaussian stripe dataset with 2D Gaussian SOS data included along the image rim. The overall, ID, and OOD accuracy of the generate images of the model trained under this dataset is shown in \textbf{(e)} along with 25 randomly selected generated images in \textbf{(f)}. }
\label{fig:comparison}
\end{figure}

In Sec.~\ref{sec:res-composition}, we have seen that the models are sufficiently data efficient in learning the compositionality given a few 2D examples. Here we compare the performance of models trained on a pure 1D Gaussian stripe dataset versus a 1D Gaussian stripe dataset + a small amount 2D compositional examples along the image rim, where the conjunction of the stripes are only partially visible within the data images. Here we name the pure 1D Gaussian stripe dataset as 1D stripes without 2D rim and the alternative as 1D stripes with 2D rim. The data latent representations of both datasets are shown in Fig.~\ref{fig:comparison}(a), (d) and the overall, in-distribution, and out-of-distribution sampled image accuracy of models trained on these datasets are shown in (b) and (e). We note from the generation behavior that the model trained without the 2D rim has a hard time generating the portion of the images where the vertical and horizontal stripes overlap. However, given a small set of examples of partial overlaps, the model has completely gained the ability to properly compose the stripes. Specifically, the model trained with the 2D rim achieves the same level of accuracy, if not better accuracy, than the 2D + 1D model shown in Fig.~\ref{fig:flag} with ample of 2D compositional examples. This suggests that models can learn compositional structure efficiently from merely a few compositional examples. Moreover, we have shown in Sec.~\ref{sec:res-composition} Fig.~\ref{fig:flag} (g) that this type of sample efficiency is transferable to an alternative type of compositionality. 

\subsection{Model's Ability to Interpolate}
\label{sec:interpolation}
\begin{figure}[h]
\begin{center}
\includegraphics[width=5.4in]{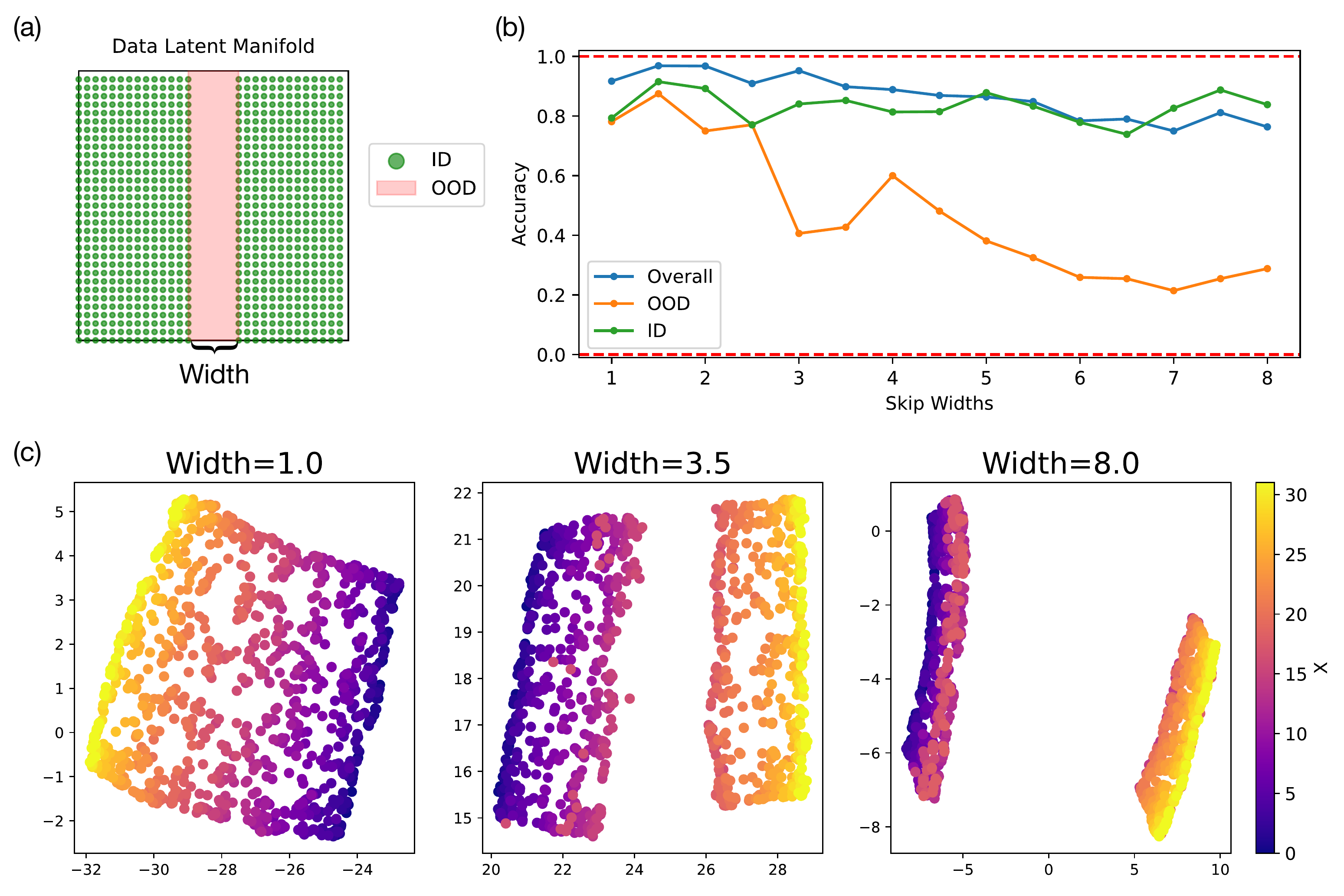}
\end{center}
\caption{\textbf{Investigation of model's ability to interpolate.} \textbf{(a)} Data manifold with a vertical lesion in the center of the manifold of various width. \textbf{(b)} Model OOD accuracy decays as a function of the width of the lesion. \textbf{(c)} The learned representation (UMAP reduced) of models trained with datasets of lesion widths 1.0, 3.5, and 8.0. }
\label{fig:interpolation}
\end{figure}

As discussed in Sec.~\ref{sec:res-composition}, the models have poor ability to interpolate as evident from the fact that models cannot generalize to unseen values of $x$ and $y$ in the OOD experiments where an entire range of $x$ and $y$ are removed from the training dataset (Fig.~\ref{fig:flag}). This is because the model learned non-parallel representations of different values of $x$'s and $y$'s, hence treating $x$ and $y$ similar to categorical values where there are non-zero overlaps between neighboring categories. We have established in Sec.~\ref{sec:res-composition} Fig.~\ref{fig:flag} that the 2D model trained on the 2D Gaussian SOS datasets with the red held-out test regions can indeed compose but not interpolate. Here we quantitatively investigate the model's ability to interpolate when trained on 2D Gaussian SOS datasets of varying widths in the held-out $x$ range, as illustrated in Fig.~\ref{fig:interpolation}(a). In Fig.~\ref{fig:interpolation}(b), we show the trained models' overall, in-distribution, and out-of-distribution image-generation accuracy as a function of the held-out range widths. We notice that the accuracy in generating images in the OOD region degrades gradually as function the held-out range width. This is likely due to the fact that the models learn parallel encodings of different values of $x$'s and $y$'s that have nontrivial overlaps with each other based on the spatial overlap of the Gaussian SOS data. Similarly the learned representations by the model become increasingly disconnected as a function of the width sizes, which is in agreement with the macroscopic generation behavior of the model.  

\subsection{Model's Ability to Compose}
\label{sec:composition}

\begin{figure}[h]
\begin{center}
\includegraphics[width=5.2in]{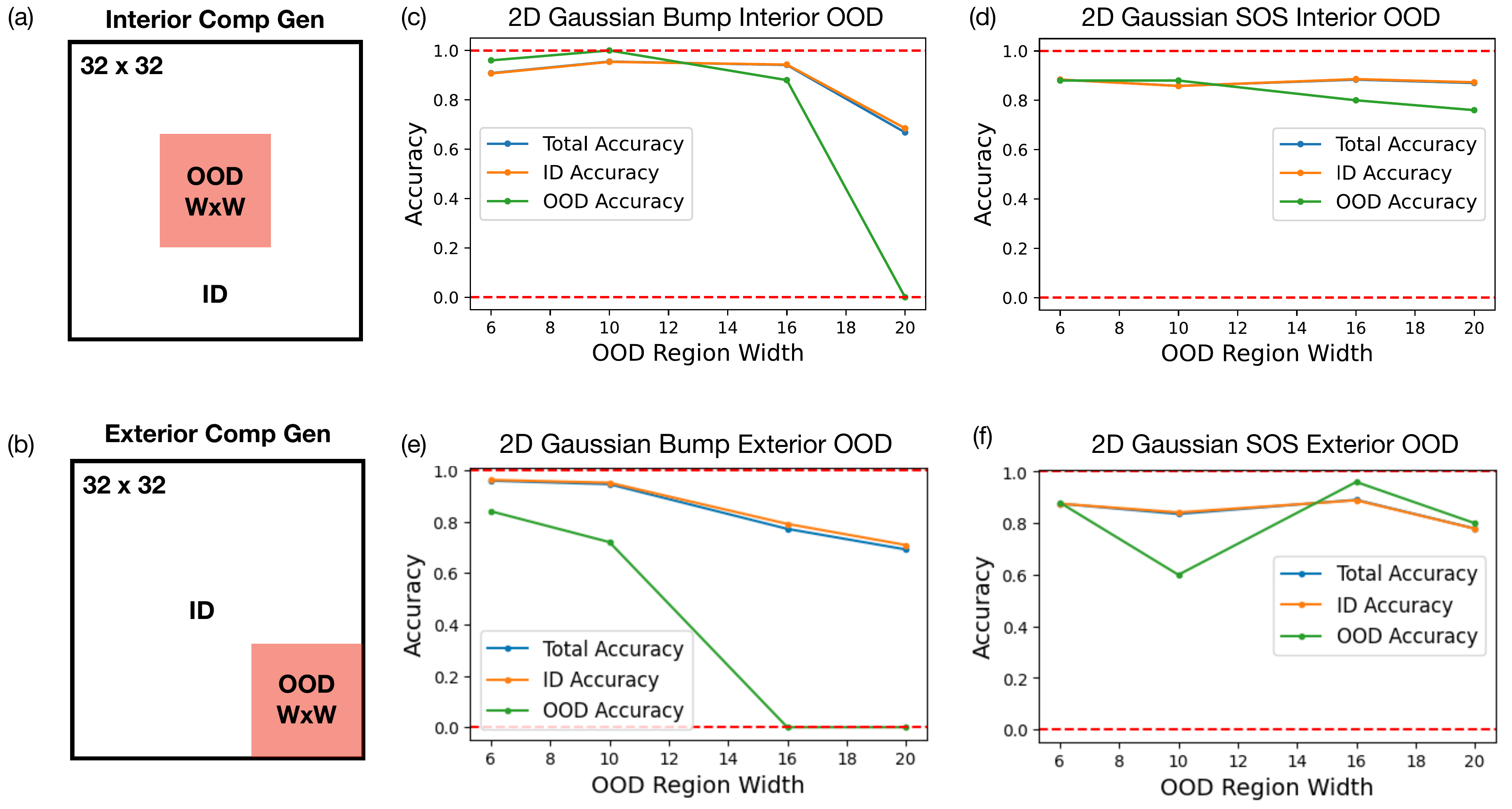}
\end{center}
\caption{\textbf{Interior and exterior compositional generalization of 2D Gaussian bump and 2D Gaussian SOS datasets.} \textbf{(a)} shows a schematic drawing of \textit{interior} compositional generalization where we leave out data points within an interior red-shaded OOD region of the data manifold. \textbf{(b)} shows the a schematic drawing of \textit{exterior} compositional generalization where we leave out data points within an exterior red-shaded OOD region of the data manifold. The size of the OOD regions in both scenario is $W\times W$. \textbf{(c),(d)} shows the accuracy of models trained on 2D Gaussian bump and 2D Gaussian SOS datasets for interior compositional generalization as a function of various OOD region width $W$. \textbf{(e),(f)} shows the accuracy of models trained on 2D Gaussian bump and 2D Gaussian SOS datasets for exterior compositional generalization as a function of various OOD region width $W$.}
\label{fig:comp_gen}
\end{figure}

In this section, we investigate model's ability to compositionally generalize when we cut out an out-of-distribution square region of size $W\times W$ in the image data manifold. Specifically, we are interested in two scenario: 1) \textit{interior compositional generalization} where the OOD cut is in the center of the data manifold, and 2) \textit{exterior compositional generalization} where the OOD cut is at the corner of the data manifold. In either case, we want to test out diffusion model's ability to generalize to the OOD regions as a function of the OOD region width $W$. We test models under both settings of interior and exterior compositional generalization when trained using a 2D Gaussian bump dataset or a 2D Gaussian SOS dataset. The results show that for the Gaussian bump dataset, the model's ability to compositionally generalize in interior regions slowly decay while in exterior regions quickly decay as a function of the OOD region width. On the other hand, the model's performance remains relatively high for all tested OOD region widths in both interior and exterior compositional generalization when trained on the Gaussian SOS dataset. These results show that the Gaussian SOS dataset indeed allows the model to acquire compositionality in a more robust as well as sample efficient manner. 

\section{Manifold formation and Percolation Theory}
\label{app-sec:percolation}
\subsection{Percolation Theoretical Simulation}
\label{sec:percolation_simulation}
\begin{figure}[h]
\begin{center}
\includegraphics[width=5.5in]{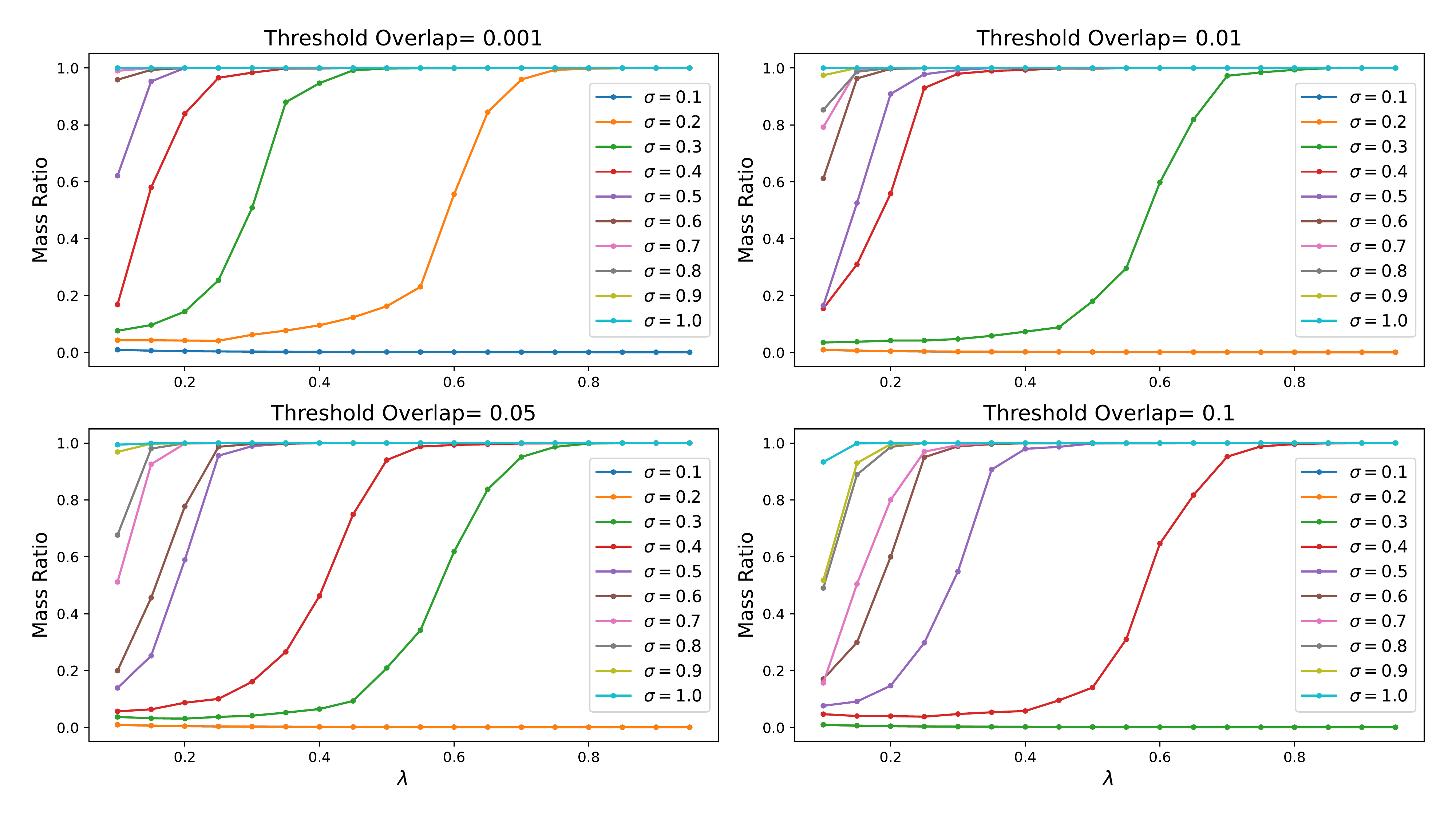}
\end{center}
\caption{\textbf{Percolation simulation of various $\sigma$'s with different overlap thresholds.} }
\label{fig:perc_simulation}
\end{figure}

In Sec.~\ref{sec:res-perc}, we have shown simulation results on percolation Fig.~\ref{fig:percolation}(c) to match with our experimental data in (e). Specifically, we perform the simulation on actual 2D Gaussian bump datasets generated in the same way as those used to train the models generating the experimental results. Here we employ a neighborhood-finding algorithm that recursively iterates through and find all island of connected clusters within a given sample of a dataset of $\sigma$ and size $\lambda \times 1024$. In the neighborhood-finding algorithm, we define a given set of two 2D Gaussian bumps to be connected if the normalized L2 metric (introduced in Appendix~\ref{sec:app_dataset} and plotted in Fig.~\ref{fig:percolation}(b)) is beyond a given threshold, which becomes a hyper-parameter in our simulation. A higher threshold overlap means that the neighborhood-finding algorithm has a more stringent standards for judging which set of bumps are connected, and that it is harder to percolate given the same amount of data. The simulation results averaged over 5 sample datasets per point are shown in Fig.~\ref{fig:perc_simulation} for various $\sigma$'s and threshold overlaps. As we increase the threshold overlap in the simulation, the percolation thresholds of datasets of various $\sigma$'s become delayed. Given a large enough threshold overlap, some of the datasets with small $\sigma$'s never percolates. Moreover, we notice that in simulation $\sigma=0.1$ never percolates due to its width of a pixel wide. Nonetheless, the model has learned to generated the $\sigma=0.1$ Gaussian bumps with a larger effective width such that the model is able to construct a faithful manifold of the $\sigma=0.1$ dataset at large $\lambda$. In Fig.~\ref{fig:percolation}(c), we have chosen to display the simulation results with threshold overlap of 0.005 to best match with the observed experimental data. Such a low threshold overlap signifies that the model is sensitive to relative spatial overlaps among data images. 

\subsection{Percolation Experimental Details}
\label{sec:percolation_exp}

\begin{figure}[h]
\begin{center}
\includegraphics[width=5.4in]{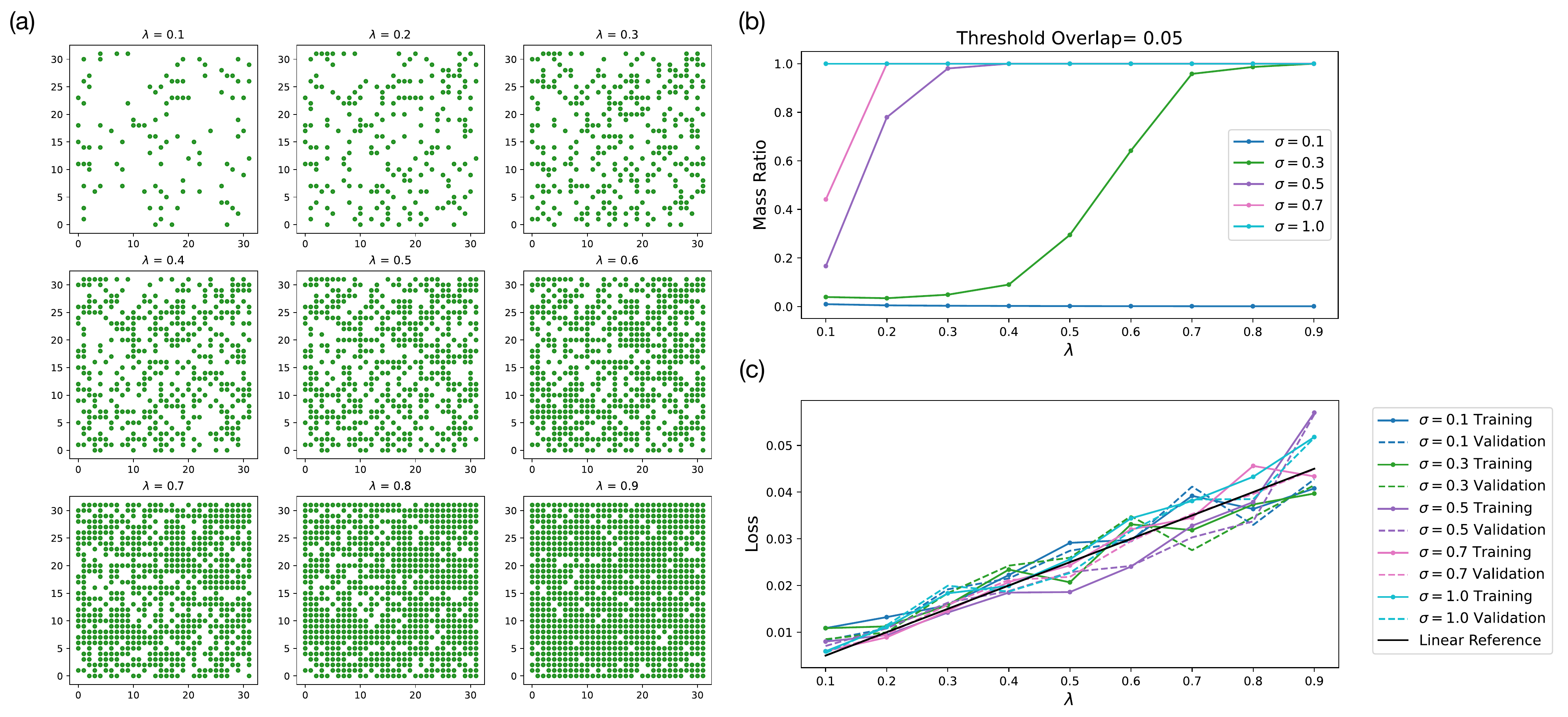}
\end{center}
\caption{\textbf{Supporting figures for percolation simulation and experiments from Fig.~\ref{fig:percolation}.} \textbf{(a)} Latent representation of datasets used to generate results in Fig.~\ref{fig:percolation}. \textbf{(b)} Simulation of largest connecting cluster mass ratio with threshold overlap of 0.05 for $\sigma$'s corresponding to those shown in Fig.~\ref{fig:percolation}(e). \textbf{(c)} The final training and validation losses as a function of $\lambda$ of models in Fig.~\ref{fig:percolation}(e). }
\label{fig:perc_exp_details}
\end{figure}

In testing out percolation theory in experiment, we relied on a specific data generation procedure to eliminate much of the stochasticity in sampling datasets to avoid having to train a large number of models. In Fig.~\ref{fig:percolation}(d), the datasets used to train model of different for different $\lambda$'s are the same across all $\sigma$'s except for the $\sigma$'s. The shared underlying latent representations of the datasets are shown in Fig.~\ref{fig:perc_exp_details}(a) for different $\lambda$'s. Moreover, we show an example simulation of the largest connected mass ratio of these specific datasets with threshold overlap of 0.05 in Fig.~\ref{fig:perc_exp_details}(b).

Note that here, as described in Sec.~\ref{sec:res-perc}, each subsequent dataset of a larger $\lambda$ is a superset of all preceding datasets of smaller $\lambda$'s. As a result, we know that if the dataset with $\sigma$ percolates at a given $\lambda_c$, then it automatically percolates at any $\lambda \geq \lambda_c$. This is an important establishment since the models trained with datasets of small $\sigma$'s seem to have high variance in the outcome training accuracy for different $\lambda$'s, as shown in Fig.~\ref{fig:percolation}(d). Given the fact that all models trained on datasets with larger $\lambda$ have a superset of information than those trained on datasets with smaller $\lambda$'s, we can attribute the variance in the experimental outcome partially to the instability in the training itself rather than percolation theory.

\begin{figure}[h]
\begin{center}
\includegraphics[width=6.0in]{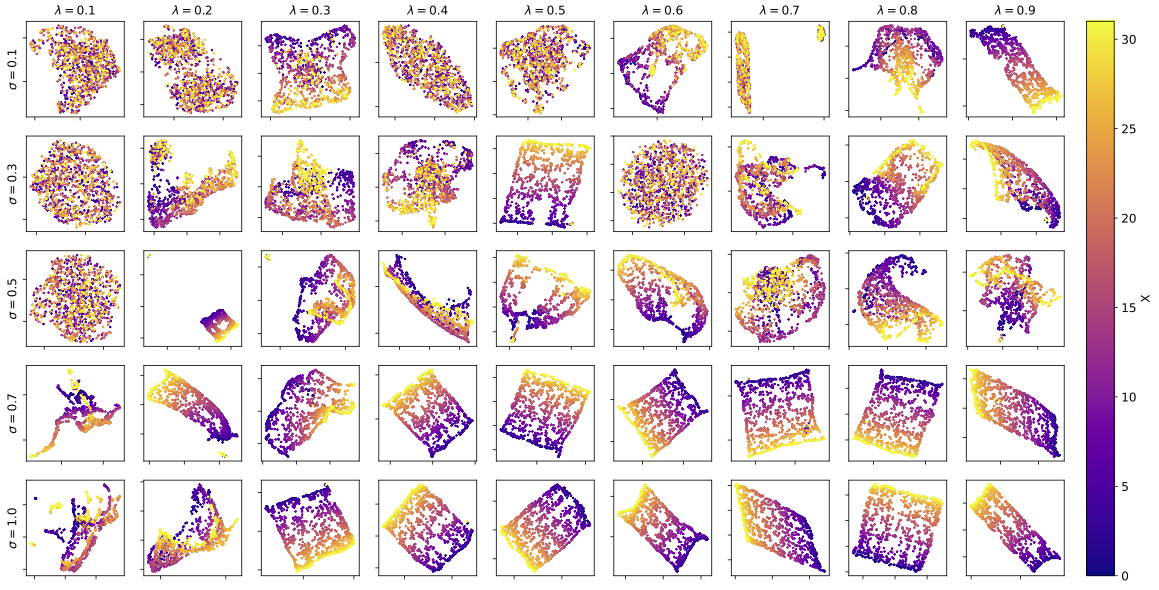}
\end{center}
\caption{\textbf{Final learned representations corresponding to models that generated results in Fig.~\ref{fig:percolation}.}}
\label{fig:perc_manifolds}
\end{figure}

One hypothesis for the observed instability in models trained on datasets with smaller $\sigma$ values is that the small interconnected components within the dataset lead to an unstable learned representation, making it more sensitive to fluctuations during optimization. The learned representation of the model corresponding to each data point shown in Fig.~\ref{fig:percolation}(d) are plotted in Fig.~\ref{fig:perc_manifolds}. As we can see, with fewer data points (smaller $\lambda$) and smaller $\sigma$, the learned representation has a less rigid structure as compared to those learned with more data and larger $\sigma$'s. Furthermore, we notice that the training and validation loss of the model grows linearly as a function of the number of training data, as shown in Fig.~\ref{fig:perc_exp_details}(c). This suggest that with a larger set of data, the model has less confidence in its learned representation due to the vast number of data it has to accommodate despite learning a higher-quality representation. Given a smaller dataset, the model can easily accommodate all data with relatively small training/validation losses despite learning a low-quality representation.     


\end{document}